\documentclass[letterpaper]{article} 
\usepackage[draft]{aaai2026}  
\usepackage{times}  
\usepackage{helvet}  
\usepackage{courier}  
\usepackage[hyphens]{url}  
\usepackage{graphicx} 
\usepackage{natbib}  
\usepackage{caption} 
\usepackage{algorithm}
\usepackage{algorithmic}

\usepackage{newfloat}
\usepackage{listings}
\DeclareCaptionStyle{ruled}{labelfont=normalfont,labelsep=colon,strut=off} 
\floatstyle{ruled}
\newfloat{listing}{tb}{lst}{}
\floatname{listing}{Listing}
\usepackage{amsmath}
\usepackage{amsfonts}  
\usepackage{adjustbox}
\usepackage[table]{xcolor}
\usepackage{tikz}
\usetikzlibrary{decorations.text}
\usetikzlibrary{calc}
\usetikzlibrary{fit}
\usetikzlibrary{bayesnet}
\usetikzlibrary{arrows}
\usetikzlibrary {arrows.meta}
\usetikzlibrary{positioning}
\usetikzlibrary{shapes}
\usetikzlibrary{shadows}
\usetikzlibrary{plotmarks}
\usepackage{booktabs} 
\usepackage{adjustbox}
\usepackage{pgfplots}
\usepackage{subfigure}
\usepackage{multirow}
\usepackage{makecell}
\usepackage{oplotsymbl}
\pgfplotsset{compat=1.18}
\newcommand{\imgbordergreen}[1]{%
  \fcolorbox{light-green}{white}{#1}%
}
\newcommand{\imgborderred}[1]{%
  \fcolorbox{light-red}{white}{#1}%
}

\definecolor{network-blue}{RGB}{165, 192, 221}
\definecolor{light-yellow}{RGB}{238, 233, 218}
\definecolor{light-yellow2}{RGB}{208, 203, 188}
\definecolor{light-green}{RGB}{129, 184, 113}
\definecolor{light-red}{RGB}{242, 182, 160}
\definecolor{light-blue}{RGB}{124, 150, 171}
\definecolor{light-orange}{RGB}{255,147,0}
\definecolor{light-purple}{RGB}{150,115,166}
\definecolor{dark-green}{RGB}{85, 124, 86}
\definecolor{dark-red}{RGB}{217, 22, 86}
\definecolor{purple}{RGB}{155, 126, 189}
\definecolor{dark-purple}{RGB}{59, 30, 84}

\title{Seeing and Knowing in the Wild: Open-domain Visual Entity Recognition with Large-scale Knowledge Graphs via Contrastive Learning}

\author{
    Hongkuan Zhou\textsuperscript{\rm 1 \rm 2}, Lavdim Halilaj\textsuperscript{\rm 1}, Sebastian Monka\textsuperscript{\rm 1}, Stefan Schmid\textsuperscript{\rm 1}, \\ Yuqicheng Zhu\textsuperscript{\rm 1 \rm 2}, Jingcheng Wu\textsuperscript{\rm 2}, Nadeem Nazer\textsuperscript{\rm 1 \rm 4}, Steffen Staab\textsuperscript{\rm 2 \rm 3}
}
\affiliations{
    \textsuperscript{\rm 1}Corporate Research, Robert Bosch GmbH, Renningen, Germany\\
    \textsuperscript{\rm 2}University of Stuttgart, Stuttgart, Germany\\
    \textsuperscript{\rm 3}University of Southampton, Southampton, UK\\
    \textsuperscript{\rm 4}Otto-von-Guericke-University Magdeburg, Magdeburg, Germany\\

    hongkuan.zhou@de.bosch.com
}

\usepackage{bibentry}

\begin{document}

\maketitle

\begin{abstract}
Open-domain visual entity recognition aims to identify and link entities depicted in images to a vast and evolving set of real-world concepts, such as those found in Wikidata. Unlike conventional classification tasks with fixed label sets, it operates under open-set conditions, where most target entities are unseen during training and exhibit long-tail distributions. 
This makes the task inherently challenging due to limited supervision, high visual ambiguity, and the need for semantic disambiguation. We propose a \textbf{Know}ledge-guided \textbf{Co}ntrastive \textbf{L}earning (KnowCoL) framework that combines both images and text descriptions into a shared semantic space grounded by structured information from Wikidata. 
By abstracting visual and textual inputs to a conceptual level, the model leverages entity descriptions, type hierarchies, and relational context to support zero-shot entity recognition.
We evaluate our approach on the OVEN benchmark, a large-scale open-domain visual recognition dataset with Wikidata IDs as the label space. Our experiments show that using visual, textual, and structured knowledge greatly improves accuracy, especially for rare and unseen entities. 
Our smallest model improves the accuracy on unseen entities by 10.5\% compared to the state-of-the-art, despite being 35$\times$ smaller.
\end{abstract}

\begin{links}
    \link{Code}{https://github.com/hk-zh/KnowCoL}
\end{links}

\maketitle

\section{Introduction}
The ability to recognize and identify visual entities in the open world is a critical milestone for scalable computer vision systems. 
In contrast to traditional classification tasks that depend on a fixed set of categories, open-world visual entity recognition seeks to recognize and link images to a vast and evolving universe of real-world entities, such as specific landmarks, artworks, biological species, or public figures. 
The recently introduced Open-domain Visual Entity recognitioN (OVEN) benchmark~\cite{OVEN} challenges models to link images and accompanying text query (specifying the intent of the image) to the correct Wikidata entities,~\footnote{We consistently use the term \textbf{entity} to refer to uniquely identifiable real-world concepts in Wikidata (e.g. Albert Einstein (Q937) and Golden Retriever (Q38686)), which differs from the typical usage of class, denoting categories within a predefined set.} identified by its unique QID.
A few examples of the OVEN-based tasks are shown in Figure~\ref{fig:motivation}a.

Previous approaches addressing open-world visual entity recognition typically adopt a dual-encoder paradigm, which aligns visual representations with textual descriptions of corresponding entities. CLIP2CLIP and CLIPFusion~\cite{OVEN} include the information of the lead image(s) of the entities into the training pipeline for a better alignment of different modalities. Recently, researchers~\cite{OVEN, GER, AutoVER} have developed \emph{two-step} approaches which leverage generative language models, such as PaLI~\cite{PaLI}, GIT~\cite{GIT}, and Vicuna~\cite{vicuna}, to generate textual labels of images and then utilize search algorithms (e.g. BM25~\cite{BM25}) to identify entities whose names closely match these predicted labels.

Despite the progress, this task remains fundamentally challenging. 
First, the label space is extremely large and long-tailed, encompassing millions of entities, many of which are rare or entirely unseen during training. 
Second, existing visual classifiers treat entities as isolated labels, ignoring the wealth of semantic relationships and factual knowledge that exist between entities. These limitations hinder generalization, especially in zero-shot settings, where the model must recognize entities it has never encountered during training.
Third, a key drawback of the two-step generative approaches is the \emph{information loss} incurred when converting rich visual content into simplified textual labels, leading to semantic ambiguity~\cite{ELSurvey, DBLP:journals/kais/BouarroudjBB22} when conducting BM25 search algorithms based on the predicted textual labels to find the final entity. Entities with similar or identical textual labels may represent fundamentally different concepts. For instance, “Mercury” can denote either the innermost planet of our Solar System or the liquid metal with chemical symbol Hg (atomic number 80), showing how simple text matching cannot distinguish these distinct meanings.

We propose that open-domain visual entity recognition should move beyond superficial recognition toward semantic-level understanding.
This involves abstracting both images and textual description into a shared conceptual space with rich structural knowledge among entities (cf. Figure~\ref{fig:motivation}b). 
\input{tikz/motivation_figure3}
In this view, recognition is not simply matching visual features to entity names, but of aligning image content with structured, contextualized knowledge about the entities depicted. 
Available knowledge graphs (KGs), such as Wikidata, which offers a rich and structured representation of hierarchical and association relations for millions of real-world entities, can be leveraged to achieve this. 

We present \textbf{Know}ledge-Guided \textbf{Co}ntrastive \textbf{L}earning (KnowCoL), an approach for open-domain visual entity recognition. 
Input images and candidate entities' descriptions and/or lead images are projected into a shared semantic embedding space, structured with the prior knowledge from the KG. 
This enables zero-shot recognition by allowing the model to generalize from seen to unseen entities based on semantic similarity, while also supporting entity disambiguation via knowledge-informed representations. 
KnowCoL is based on a dual-encoder paradigm, eliminating the information loss in the process of conversion in the two-step approaches.
We evaluate our approach on the OVEN benchmark. 
Our contributions are as follows:
\begin{enumerate}
    \item We present a dual-encoder approach for the open-domain visual entity recognition task that leverages external knowledge from both Wikidata and Wikipedia to enable a better representation of images and text descriptions.
    \item We investigate the impact of different forms of external knowledge (including lead images) and different relations between entities on the recognition performance.
    \item Our approach demonstrates strong zero-shot generalization on the OVEN benchmark, showing that incorporating external knowledge significantly improves entity recognition performance of unseen entities by 10.5\%, even with models that are 35$\times$ smaller.
\end{enumerate}

\section{Preliminary}
\subsection{Contrastive Loss}
Contrastive loss is a widely used objective in representation learning, particularly in contexts such as metric learning, image-text alignment, and self-supervised learning. Given two sets of embeddings $A = \{a_i\}_{i=1}^N$ and $B = \{b_i\}_{i=1}^N$, where $(a_i, b_i)$ is a positive (matching) pair and $\forall (a_i, b_j), j\neq i$, is a negative pair, we define the contrastive loss  as:

\begin{equation}
\ell(A, B) = -\frac{1}{N} \sum_{i=1}^{N} \log \frac{\exp\left( \text{sim}(a_i, b_i) / \tau \right)}{\sum_{j=1}^{N} \exp\left( \text{sim}(a_i, b_j) / \tau \right)} \text{,}
\label{eq:contrastive loss function}
\end{equation}
where $\text{sim}(\cdot,\cdot)$ denotes the similarity function, $\tau$ is a temperature parameter controlling the softness of the similarity distribution, and $N$ is the set size. Common similarity functions include cosine similarity and Euclidean distance-based similarity. 
Specifically, the symmetric contrastive loss can be defined as, 
\begin{equation}
    \ell_\text{sym} (A,B) = \frac{1}{2}\left(\ell\left(A,B\right) + \ell\left(B,A\right)\right)\text{,}
    \label{eq:symmetric contrastive loss function}
\end{equation}
which encourages mutual alignment, and it is widely used in various models like CLIP. 

\subsection{Wikidata Knowledge Graph}

Wikidata is a collaboratively-maintained, multilingual knowledge graph that can be formalised as the directed graph $\mathcal{G}=(\mathcal{E},\mathcal{R},\mathcal{T})$, where $\mathcal{E}$ is the set of entities (called items and uniquely referenced by ``Q-identifiers'' (QIDs), e.g., Q42 for Douglas Adams), $\mathcal{R}$ is the set of relation types (called properties and referenced by ''P-identifiers`` (PIDs), e.g., P31 for instance of), and $\mathcal{T}\subseteq\mathcal{E}\times\mathcal{R}\times\mathcal{E}$ is the set of knowledge triples $(e_1,r,e_2)$ stating that subject entity $e_1$ is linked to object entity $e_2$ via property $r$. As of 2025, Wikidata KG contains more than 100 million entities, 12,000 properties, and 10 billion triples and is growing daily through human and bot contributions.

\subsection{Wikipedia Knowledge Base}
Wikipedia is a collaboratively edited, multilingual encyclopedia whose every article is linked to a unique Wikidata QID. This tight coupling lets each page serve as a multimodal hub: rich encyclopedic text is paired with lead images that provide both descriptive and visual context for the corresponding entity in the Wikidata KG.

\section{Related Work}

\subsection{Zero-Shot Classification}
Open-world recognition is closely related to zero-shot learning (ZSL), where models must recognize classes with no training images by relying on auxiliary information. 

Early approaches~\cite{DBLP:conf/cvpr/FarhadiEHF09, DBLP:conf/cvpr/LampertNH09} introduced human-defined attributes as intermediate semantic representations for recognition of unseen classes. 
Later methods moved toward unsupervised semantic embeddings derived from text corpora~\cite{DBLP:conf/nips/SocherGMN13, DBLP:conf/nips/FromeCSBDRM13, DBLP:journals/corr/NorouziMBSSFCD13}. These dual-encoder or projection-based approaches have proven that aligning images with distributed text representations enables zero-shot recognition at scale.  
Vision-language pretraining has greatly advanced this paradigm. A well-known example is CLIP~\cite{CLIP}, which is trained on 400 million image-text pairs to learn a joint embedding space for images and natural language. By encoding an image and a candidate label (or description) into the same space, CLIP can directly measure their compatibility, essentially performing zero-shot classification by ranking labels. CLIP has wide applications in visual inspection~\cite{jeong2023winclip, MultiADS2025} and robotics~\cite{zhou2023language, 10685120, 10934975}. 

Another line of zero-shot research uses large-scale generative models to bridge the gap to unseen classes.  Instead of directly mapping images to text semantics, these generative approaches output the entity names based on given images and then link the entity names to the categories (two-step approach). For instance, \citet{OVEN} use PaLI and BLIP-v2~\cite{blipv2} on the OVEN benchmark to make the model predict the entity name and then uses the BM25 search algorithm to find the closest label in the label space of Wikidata ids. Instead of generating entity names, GER-ALD~\cite{GER} generates the semantic and discriminative ``code'' to identify entities. Auto-VER~\cite{AutoVER} combines contrastive learning with generative models to enhance the ability to distinguish similar entities within a vast label space. Unlike dual-encoder approaches, two-step generative methods output text conditioned on images and then match them to labels in the label space. This introduces semantic ambiguity, as generated outputs may be correct lexically but vague or incomplete semantically.

\subsection{Knowledge Graphs for Visual Entity Recognition}
Many studies have explored incorporating knowledge into vision-and-language tasks~\cite{DBLP:journals/semweb/MonkaHR22}, including visual question answering~\cite{DBLP:conf/cvpr/ChangCNGSB22, DBLP:conf/semweb/0007CGPYC21} and entity-aware image captioning~\cite{DBLP:conf/cvpr/BitenGRK19}. For visual entity recognition task, \citet{DBLP:conf/cvpr/0004YG18} and \citet{DBLP:conf/cvpr/KampffmeyerCLWZ19} use semantic embeddings and categorical relationships from KGs through graph convolution networks for zero-shot prediction. \citet{10038499} augments few-shot image recognition by introducing auxiliary semantic prior knowledge and propagating knowledge among categories via a semantic-visual mapping.
KG-NN~\cite{DBLP:conf/semweb/MonkaH0R21} combines prior knowledge encoded in KGs with visual representations to enhance generalization under distribution shifts, and KGV~\cite{DBLP:journals/corr/abs-2410-15981} demonstrates that the integration of richer multi-modal priors can further improve performance. Instead of relying on small domain-specific knowledge graphs, our approach is the first to leverage a large-scale knowledge graph for open-domain visual entity recognition, demonstrating strong scalability potential.

\section{Methodology}
\input{tikz/architecture}
Our goal is to integrate the rich, structured knowledge from Wikidata into the training pipeline. To achieve this, we align knowledge graph embeddings (KGEs), image embeddings, and text embeddings within a shared latent space, enabling the structured knowledge captured by the KGEs to be effectively injected into the latent space's representation. A framework overview can be seen in Figure~\ref{fig:architecture}.

\subsection{Problem Definition}
Let $\mathcal{E}$ be the set of Wikidata entities identified by QIDs.  Define the input space as $\mathcal{X}= \mathcal{I}\!\times\!\mathcal{L}$, where $\mathcal{I}$ denotes the domain of RGB images and $\mathcal{L}$ is the linguistic (text) space.  The Open-domain Visual Entity Recognition (OVEN) task seeks a function $f:\mathcal{X}\rightarrow\mathcal{E}$ which maps each image-text pair ($x^p$, $x^t$) to the unique entity $e \in \mathcal{E}$. Every entity $e$ is associated with a Wikipedia-derived record $(t_e,\,I_e)\in\mathcal{K}\subseteq\mathcal{L}\times2^{\mathcal{I}}$, where $t_e\in\mathcal{L}$ is its encyclopaedic description and $I_e\subset\mathcal{I}$ is the finite set of lead images illustrating the corresponding entity. Given a labeled training dataset $\{(x_i,e_i)\}_{i=1}^N\subset\mathcal{X}\times\mathcal{E}$, the goal is to learn a function $f$ that minimizes the number of misclassifications.


\subsection{Embeddings}
\subsubsection{Knowledge Graph Embeddings}
To model the knowledge existing in the knowledge graph $\mathcal{G}$, we adopt embedding methods to inject the knowledge into the latent space.
We define knowledge graph embeddings as mappings:
\[
    \phi: \mathcal{E} \to \mathcal{Z}, \quad \psi: \mathcal{R} \to \mathbb{R}^{d_r} \text{,}
\]
where $\mathcal{Z} \subseteq \mathbb{R}^{d_e}$ is the latent space with dimension $d_e$. $\phi(e), \psi(r)$ are  the node and relation embeddings, respectively. Common knowledge graph embedding methods define score function $f_s(e_1,r,e_2)$ to measure the plausibility of a triple $(e_1,r,e_2) \in \mathcal{T}$.

\subsubsection{Image Embeddings}
Let $\mathcal{I}$ be the space of input images. We define an image mapping function as:
\[
    f_{\mathbf{\theta}}: \mathcal{I} \to \mathcal{Z}\text{,}
\]
where $f_{\mathbf{\theta}}$ is represented by a deep neural network with parameters $\mathbf{\theta}$. 
Here, we leverage the pretrained frozen CLIP image encoder plus a trainable linear project layer to be $f_{\mathbf{\theta}}$. 

\subsubsection{Text Embeddings}
Let $\mathcal{L}$ be the linguistic (text) space of input text descriptions. We define a text mapping function:
\[
    f_{\mathbf{\lambda}}: \mathcal{L} \to \mathcal{Z}\text{,}
\]
where $f_{\mathbf{\lambda}}$ is represented by a deep neural network with parameters $\mathbf{\lambda}$. We leverage the pretrained \emph{frozen} CLIP text encoder plus a \emph{trainable} linear project layer to be $f_{\mathbf{\lambda}}$.

We employ a dual-encoder approach to align the semantic representation of the input image and text query with that of the corresponding entity, which includes its lead image(s) and textual description. We structure our discussion into two parts: (1) encoding the input image and text query, and (2) encoding the multi-modal information of entities from the knowledge graph.
\subsubsection{Input Image and Text Query Encoding}
To fuse the information of the input image $x^p$ and the text query $x^t$, we employ an extra encoder
\[
   f_{\mathbf{\gamma}}: (\mathcal{Z}, \mathcal{Z}) \to \mathcal{Z}
\]
with parameters $\mathbf{\gamma}$ to fuse the image embedding $f_{\mathbf{\theta}}(x^p)$ and text embedding $f_{\mathbf{\lambda}}(x^t)$. The embedding, combining the information from the input image and text query, can be written as 
\begin{equation}
    \mathbf{z}^\text{input} = f_{\mathbf{\gamma}}(f_{\mathbf{\theta}}(x^p), f_{\mathbf{\lambda}}(x^t))
    \label{equ: input_embedding}
\end{equation}

\subsubsection{Entity's Multi-modal Information Encoding}
For each entity $e$, $t_e$ and $I_e$ represent its text description and set of lead image(s), respectively. The entity text embedding can be written as
\begin{equation}
        \mathbf{z}^\text{entityText} = f_\lambda(t_e)\text{,}
        \label{eq: entity text embedding}
\end{equation}
and the image embedding can be written as
\begin{equation}
    \mathbf{z}^\text{entityImage} =
    \begin{cases}
     \frac{1}{|I_e|}\sum_{a \in I_e} f_\theta(a)  & \text{if } I_e \ne \emptyset \\
     \mathbf{z}^\text{entityText} & \text{if } I_e = \emptyset 
    \end{cases} \text{.}
    \label{eq: entity image embedding}
\end{equation}
If no lead image is available for a given entity, we assign $\mathbf{z}^\text{entityText}$ to $\mathbf{z}^\text{entityImage}$. This ensures a smooth fusion of visual and textual information, as discussed in the next section.

\subsection{Knowledge-guided Contrastive Learning}
For given input batch $ \{(x^p_i, x^t_i)\}_{i=1}^{N_b}$, where $x^p_i$ and $x^t_i$ are the $i$-th input images and its corresponding text query, $N_b$ is the batch size, the corresponding ground truth label set is denoted as $\{e_i\}_{i=1}^{N_b}$.   We define three losses, namely alignment loss, proxy loss, and knowledge graph loss, for the training process. Figure~\ref{fig:latent_space} shows the learning visualization. 

\subsubsection{Alignment Loss}
The alignment loss aims to align the joint representation of input images and text queries $\{\mathbf{z}^\text{input}_i\}_{i=1}^{N_b}$ with their corresponding entity representations $\{\phi(e_i)\}_{i=1}^{N_b}$. Specifically, we define the alignment loss using a contrastive objective that pulls matched embeddings closer together while pushing unmatched embeddings among the batch apart. 
It can be defined as 
\begin{equation}
\mathcal{L}_a = 
\ell_\text{sym}\left( \{\mathbf{z}^\text{input}_i\}_{i=1}^{N_b}, \{\phi(e_i)\}_{i=1}^{N_b}\right)
\text{,}
\end{equation}
where $\ell_\text{sym}(\cdot,\cdot)$ represents the contrastive loss function define in Equation~\ref{eq:symmetric contrastive loss function}.

\subsubsection{Proxy Loss}
The proxy loss is defined to align the node embeddings $\{\phi(e_i)\}_{i=1}^{N_b}$ with their corresponding multi-modal representations $\{\mathbf{z}^\text{entity}_i\}_{i=1}^{N_b}$. This objective ensures that the node embeddings capture semantic information from both image and text modalities. It can be defined as:
\begin{equation}
\begin{split}
\mathcal{L}_p =  
\frac{1}{2}\ell_\text{sym}\left(\{\phi(e_i)\}_{i=1}^{N_b}, \{\mathbf{z}^\text{entityText}_i\}_{i=1}^{N_b}\right)+ \\ \frac{1}{2}\ell_\text{sym}\left(\{\phi(e_i)\}_{i=1}^{N_b}, \{\mathbf{z}^\text{entityImage}_i\}_{i=1}^{N_b}\right) 
\end{split}
\end{equation}

We refer to this as the proxy loss because the node embeddings serve as proxies for capturing the multi-modal semantic information of the entities. Figure~\ref{fig:latent_space} shows that the proxy loss gathers image and text embeddings with their corresponding node embeddings. 

\input{tikz/latent_space2}
\subsubsection{Knowledge Graph Embedding Loss (KE Loss)}
The knowledge graph loss aims to capture the structural knowledge associated with entity $e_i$. We first extract the triplet set associated with the entity $e_i$ by $\mathcal{T}_{e_i} = \{(h,r,t) \in \mathcal{T}\mid h=e_i \lor t = e_i\}$. For each triplet $(h,r,t)\in \mathcal{T}_{e_i}$, we define $\mathcal{T}_{(h,r,t)}$ as the set of negative samples generated by corrupting either the head or tail entity in the positive triplet:
\begin{equation}
    \mathcal{T}_\text{(h,r,t)} = \{(h^\prime,r,t)\mid h^\prime \in \mathcal{E} \setminus \{h\} \} \cup \{(h,r,t^\prime)\mid t^\prime \in \mathcal{E} \setminus \{t\} \} \text{.}
    \label{equ:nagative sampling}
\end{equation}
Based on that, the knowledge graph loss can be defined as

\begin{equation}
\begin{aligned}
\mathcal{L}_{\text{KE}}
  &= \sum_{i=1}^{N_b} \frac{1}{|\mathcal{T}_{e_i}|}
     \sum_{(h,r,t)\in\mathcal{T}_{e_i}}
     \frac{1}{|\mathcal{T}_{(h,r,t)}|} \\ 
     & \quad \left[-\log\frac{\exp(f_s(h,r,t)/\tau)}{
     \sum_{(h',r,t')\in\mathcal{T}_{(h,r,t)}}\exp(f_s(h^\prime,r,t^\prime)/\tau)}\right]\text{,}
\end{aligned}
\label{eq:ke-loss}
\end{equation}
where $\tau$ is a temperature and $f_s(\cdot)$ is the score function defined by
\begin{equation}
    f_s(h,r,t) = \frac{(\phi(h) + \psi(r))^\top \phi(t)}{\|\phi(h) + \psi(r)\|_2 \, \|\phi(t)\|_2} \text{,}
\end{equation}
which is the cosine similarity between $\phi(h)+\psi(r)$ and $\phi(t)$. Here, we follow a translation-based KGE approach with a cosine similarity distance metric. 
The final loss for optimization can be defined as:
\begin{equation}
    \mathcal{L} = \mathcal{L}_a + \beta_1\mathcal{L}_p + \beta_2\mathcal{L}_\text{KE} \text{,}
\end{equation}
where $\beta_1$ and $\beta_2$ are two hyperparameters for balancing. 

\subsection{Inference}
Given an input image $x^p$ and a text query $x^t$, we first obtain the input embedding $\mathbf{z}^{\text{input}}$ as defined in Equation~\ref{equ: input_embedding}. The corresponding entity answer $e^*$ is then identified through the following inference step:
\begin{equation}
e^* = \arg\max_{e_j \in \mathcal{E}} \text{sim}\left(\mathbf{z}^{\text{input}},\,\frac{1}{2} \left( \mathbf{z}^{\text{entityText}}_j + \mathbf{z}^{\text{entityImage}}_j\right) \right) \text{,}
\end{equation}
where $\mathbf{z}^{\text{entityText}}_j$ and $\mathbf{z}^{\text{entityImage}}_j$ denote the textual and visual embeddings of entity $e_j$, respectively, as defined in Equations~\ref{eq: entity text embedding} and \ref{eq: entity image embedding}.

\section{Experiments}
In our experiments, we investigate whether incorporating knowledge from large-scale KGs and Knowledge Base (KB) can enhance open-domain visual entity recognition. We analyze the impact of different types of prior knowledge, fusion methods, and KGE techniques. 
In addition, ablation studies on various hyperparameter values are also provided.


\subsection{Experiment Settings}
\begin{table*}[htbp]
    \centering
    \begin{tabular}{l | l c c c c c | c |}
        \toprule
            Paradigm & Model & Venues & Parameters (B) & Pre-train Dataset & Seen & Unseen & HM \\
        \midrule
            \multirow{3}{*}{\makecell{Dual-Encoder }}& CLIP~\cite{CLIP} & ICML2021 & 0.4 & OpenAI & 5.6 & 4.9 & 5.2\\
            & CLIPFusion~\cite{OVEN} & ICCV2023 & 0.9 & OpenAI & 33.6 & 4.8 & 8.4 \\
            & CLIP2CLIP~\cite{OVEN} & ICCV2023 & 0.9 & OpenAI & 12.6 & 10.5 & 11.4 \\
            \midrule
            \multirow{7}{*}{\makecell{Two-Step \\ Generative}} & BLIP-v2~\cite{blipv2} & ICML2023 & 12.2 & - & 8.6 & 3.4 & 4.9 \\
            & PaLI-3B~\cite{OVEN} & ICCV2023 & 3.0 & WebLI & 19.1 & 6.0 & 9.1 \\
            & PaLI-17B~\cite{OVEN} & ICCV2023 & 17.0 & WebLI & 28.3 & 11.2 & 16.0 \\
            & GIT-Large~\cite{GER} & CVPR2024 & 0.4 & WebLI & 17.6 & 4.3 & 7.0\\
            & GER-ALD~\cite{GER} & CVPR2024 & 0.4 & LAION & 29.1 & 16.3 & 20.9\\
            & Auto-VER-7B~\cite{AutoVER} & ECCV2024 & 7.0 & - & 62.8 & 16.0 & 25.5\\
            & Auto-VER-14B~\cite{AutoVER} & ECCV2024 & 14.0 & - & \textbf{65.0} & 18.6 & 28.9 \\
        \midrule
            \makecell{Knowledge-Guided\\  Dual-Encoder} & KnowCoL-bigG(ours) & - & 2.0 & LAION & 41.8 & \textbf{36.1} & \textbf{38.8}\\
        \bottomrule
    \end{tabular}
    \caption{Comparison with current state-of-the-art approaches on OVEN entity test split. We evaluate the harmonic mean (HM) of the seen and unseen splits (top1 accuracy) after fine tuning on OVEN training set. It serves as our main metric. The numbers of baselines are taken from papers~\cite{GER, OVEN, AutoVER}. The total parameters and pre-training datasets of the models are given for comparison.}
    \label{tab:Comparison with the state of the art}
\end{table*}
\subsubsection{OVEN Dataset}
The OVEN dataset contains 6,063,945 training samples, which combines 14 image recognition datasets. 
Models are evaluated on the test splits, and performance is reported using the harmonic mean (HM) of top-1 accuracy for two types of entities: seen entities, which appear in the OVEN training set, and unseen entities, which are not present during training. Specifically, the test split includes 15,888 entities (8,355 seen and 7,533 unseen). 
Our evaluation is conducted solely on the OVEN benchmark, comprising 14 diverse image recognition datasets, thus indicating the generalizability of our approach. 
The evaluation metric is top-1 accuracy for seen and unseen entities and the HM of them. 
\subsubsection{Training details}
We extract a subgraph from Wikidata containing 32,122 entities and 501 relation types. 
For training, we use the AdamW optimizer with a learning rate of 0.001, batch size of 4096, and weight decay of 0.0001. 
Our approach includes three model variants based on different OpenCLIP image encoders: ViT-L/14, ViT-H/14, and ViT-bigG/14. The temperature $\tau$ for contrastive learning is fixed at 0.07 during fine-tuning. The hyperparameters $\beta_1$ and $\beta_2$ are both set to 1. We fuse information from the input image and text query (Equation~\eqref{equ: input_embedding}), using a simple \emph{addition} operation. More details are in the Appendix.

\subsection{Baselines}
\subsubsection{Dual Encoders}
Dual-encoder approaches consist of two encoders, each encoding a dedicated modality (e.g., images and text) into the \emph{same} latent space. 
The visual recognition is conducted by searching the nearest neighbors of prototypes in the latent space for a given input image. 
Representative models include CLIP, CLIPfusion, and CLIP2CLIP. 

\subsubsection{Auto-regressive Captioning}
Another approach uses auto-regressive models to generate captions for the given images. Representative vision-language models (VLMs) such as PaLI, GIT-Large, GER-ALD, and Auto-VER follow this paradigm. These models generate a descriptive caption or name for the visual entity and then match it to the closest label in the predefined label space for final prediction. 

\subsubsection{Comparison with the state of the art}
We compare our proposed KnowCoL model with both dual-encoder and auto-regressive captioning baselines. As shown in Table~\ref{tab:Comparison with the state of the art}, our model achieves significantly stronger performance on both seen and unseen entities. KnowCoL-bigG achieves the highest harmonic mean (HM) of 38.8\%, outperforming large-scale generative models such as Auto-VER-14B and PaLI-17B despite using 7$\times$ fewer parameters. We identify that the Auto-VER achieves impressive results for seen entities, but the performance significantly drops for unseen entities, indicating overfitting. Compared to theirs, KnowCoL achieves a balanced result on both seen and unseen entities, representing a strong generalization ability by including structure, textual, and visual prior knowledge.





\subsection{Analysis and Ablation Studies}
Here, we analyse the impact of model size, prior knowledge types, fusion strategy of the text query and input image, various KGE methods, and hyperparameters - $\beta_1$, $\beta_2$, latent space dimension $d_e$, and temperature $\tau$. The detailed setting can be found in the Appendix section, experiment setting. 
\subsubsection{Impact of Model Size}
\begin{table}[htbp]
    \centering
    \begin{tabular}{l l c c c |c|}
        \toprule
            Model & Backbone & Para.(B) & Seen & Unseen & HM \\
        \midrule
            CLIP & ViT-L-14 & 0.4 & 5.6 & 4.9 & 5.2 \\
            CLIPFusion & ViT-L-14 & 0.9 & 33.6 & 4.8 & 8.4 \\
            CLIP2CLIP & ViT-L-14 & 0.9 & 12.6 & 10.5 & 11.4 \\  
            KnowCoL-L & ViT-L-14 & 0.4 & \textbf{34.3} & \textbf{29.1} & \textbf{31.5}\\
            \midrule
            KnowCoL-H  & ViT-H-14 & 0.9 & 38.5 & 33.4 & 35.8\\
            KnowCoL-bigG & ViT-bigG-14 & 2.0& \textbf{41.8} & \textbf{36.1} & \textbf{38.8} \\
        \bottomrule 
    \end{tabular}
    \caption{Comparison of model sizes. ViT-H-14 and ViT-bigG-14 are used for our KnowCoL-H and KnowCoL-bigG models. ViT-L-14 is used for CLIP, CLIPFusion, CLIP2CLIP, and KnowCoL-L. Note that CLIPFusion and CLIP2CLIP utilize two ViT-L-14 backbones.}
    \label{fig:model size}
\end{table}
As Table~\ref{fig:model size} demonstrates, we evaluate our model with different CLIP backbones. The backbones ViT-L-14, ViT-H-14, and ViT-bigG-14 are used for KnowCoL-L, KnowCoL-H, and KnowCoL-bigG, respectively. 
The performance across both seen and unseen entities consistently improves as model size increases. 

We highlight that KnowCoL-L, CLIP, CLIPFusion, and CLIP2CLIP use the same backbone ViT-L-14, but KnowCoL-L significantly outperforms all. While CLIPFusion and CLIP2CLIP achieve harmonic means of 8.4\% and 11.4\%, respectively, KnowCoL-L reaches 31.5\%, showing about 2.8 times improvement. This demonstrates that external knowledge significantly improves zero-shot generalization without increasing model size. 

\begin{figure*}[htbp]
  \centering
  \subfigure[Hyperparameter $\beta_1$]{
    \begin{tikzpicture}
      \begin{axis}[
        ylabel={HM},
        ylabel near ticks,
        ymin=27.5,           
        ymax=44,
        width=4.5cm,
        height=4cm,
        grid=major,
        xtick={0.1, 0.5, 1},
        xticklabels={$0.1$, $0.5$, $1$},
        legend columns=1,
        legend style={
          at={(1,1)},
          anchor=north east,
          font=\scriptsize,
        }
        ]
        \addplot[color=light-blue, mark=*] coordinates {
          (0.1,29.5) (0.5,31.1) (1.0,31.5)
        };
        \addlegendentry{KnowCoL-L}
        \addplot[color=light-purple, mark=*] coordinates {
          (0.1,35) (0.5,36.4) (1.0,36.9)
        };
        \addlegendentry{KnowCoL-H}
        
        \addplot[
            color=light-blue,
            mark=pentagon*,
            only marks,
            mark options={fill=light-blue, scale=1.5}
        ] coordinates {
        (1.0,31.5)
        };
        \addplot[
            color=light-purple,
            mark=pentagon*,
            only marks,
            mark options={fill=light-purple, scale=1.5}
        ] coordinates {
        (1.0,36.9)
        };
      \end{axis}
    \end{tikzpicture}
  }
  \hfill
  \subfigure[Hyperparameter $\beta_2$]{
    \begin{tikzpicture}
      \begin{axis}[
            ylabel style={font=\tiny},
            ylabel={HM},
            xlabel near ticks,
            ylabel near ticks,
            ymin=27.5,           
            ymax=44,
            width=4.5cm,
            height=4cm,
            grid=major,
            xtick={0.1, 0.5, 1},
            xticklabels={$0.1$, $0.5$, $1$},
            legend columns=1,
            legend style={
                at={(1,1)},
                anchor=north east,
                font=\scriptsize,
                row sep=0.1pt,               
                inner sep=0pt,             
            }
        ]
        \addplot[color=light-blue, mark=*] coordinates {
          (0.1, 29.8) (0.5, 30.4) (1.0,31.5)
        };
        \addlegendentry{KnowCoL-L}
        \addplot[color=light-purple, mark=*] coordinates {
          (0.1, 34.0) (0.5, 34.5) (1.0,36.9)
        };
        \addlegendentry{KnowCoL-H}

        \addplot[
            color=light-blue,
            mark=pentagon*,
            only marks,
            mark options={fill=light-blue, scale=1.5}
        ] coordinates {
        (1.0, 31.5)
        };
        \addplot[
            color=light-purple,
            mark=pentagon*,
            only marks,
            mark options={fill=light-purple, scale=1.5}
        ] coordinates {
        (1.0,36.9)
        };
      \end{axis}
    \end{tikzpicture}
  }
  \hfill
  \subfigure[Latent Space Dim $d_e$]{
    \begin{tikzpicture}
      \begin{axis}[
        ylabel={HM},
        ylabel near ticks,
        ymin=27.5,           
        ymax=44,
        width=4.5cm,
        height=4cm,
        grid=major,
        xtick={512, 768, 1024},
        xticklabels={$512$, $768$, $1024$},
        legend columns=1,
        legend style={
          at={(1,1)},
          anchor=north east,
          font=\scriptsize,
        }
        ]
        \addplot[color=light-blue, mark=*] coordinates {
          (512,30.9) (768,31.5) (1024,31.2)
        };
        \addlegendentry{KnowCoL-L}
        \addplot[color=light-purple, mark=*] coordinates {
          (512,34.5) (768,35.3) (1024,35.8)
        };
        \addlegendentry{KnowCoL-H}
        
        \addplot[
            color=light-blue,
            mark=pentagon*,
            only marks,
            mark options={fill=light-blue, scale=1.5}
        ] coordinates {
        (768,31.5)
        };
        \addplot[
            color=light-purple,
            mark=pentagon*,
            only marks,
            mark options={fill=light-purple, scale=1.5}
        ] coordinates {
        (1024,35.8)
        };
      \end{axis}
    \end{tikzpicture}
  }
  \hfill
  \subfigure[Temperature $\tau$]{
    \begin{tikzpicture}
      \begin{axis}[
        ylabel={HM},
        ylabel near ticks,
        ymin=27.5,           
        ymax=44,
        width=4.5cm,
        height=4cm,
        grid=major,
        xtick={0.05, 0.07, 0.1},
        xticklabels={$0.05$, $0.07$, $0.1$},
        legend columns=1,
        legend style={
          at={(1,1)},
          anchor=north east,
          font=\scriptsize,
        }
        ]
        \addplot[color=light-blue, mark=*] coordinates {
          (0.05,29.7) (0.07,31.5) (0.1,30.3)
        };
        \addlegendentry{KnowCoL-L}
        \addplot[color=light-purple, mark=*] coordinates {
          (0.05,34.9) (0.07,35.8) (0.1,35.2)
        };
        \addlegendentry{KnowCoL-H}
        
        \addplot[
            color=light-blue,
            mark=pentagon*,
            only marks,
            mark options={fill=light-blue, scale=1.5}
        ] coordinates {
        (0.07,31.5)
        };
        \addplot[
            color=light-purple,
            mark=pentagon*,
            only marks,
            mark options={fill=light-purple, scale=1.5}
        ] coordinates {
        (0.07,35.8)
        };
      \end{axis}
    \end{tikzpicture}
  }
  \caption{Ablation Studies for hyperparameter $\beta_1$, $\beta_2$, latent space dimension $d_e$, and temperature $\tau$. HM indicates the harmonic mean of accuracies of seen and unseen entities. \pentagofill - represents the default setting of KnowCoL approach.}
  \label{fig:Hyperparameters}
\end{figure*}
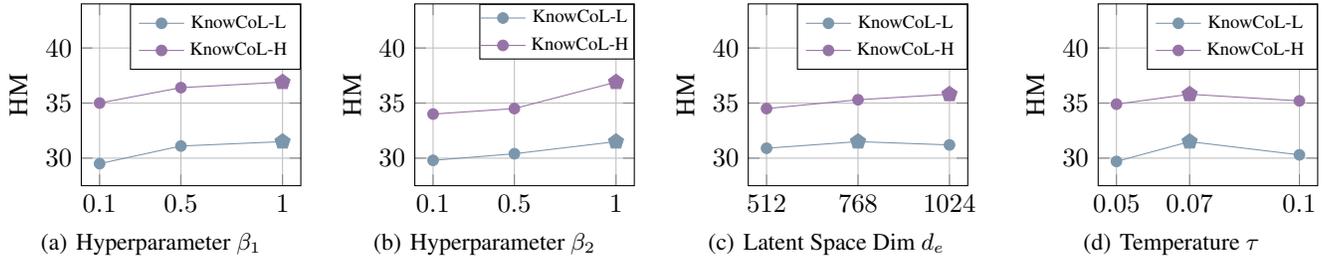
\subsubsection{Comparison of Knowledge Types}
\begin{table}[htbp]
    \centering
    \resizebox{0.98\linewidth}{!}{
    \begin{tabular}{l c c |c|}
        \toprule
            Model & Seen & Unseen & HM \\
        \midrule
            \rowcolor{gray!30} KnowCoL-L (ViT-L-14) & \textbf{34.3} & \textbf{29.1} & \textbf{31.5}\\
            \midrule
            w/o Lead images & 32.5 & 27.9 & 30.0 \\
            w/o KG Hierarchical Knowledge  & 32.1 & 25.3 & 28.3 \\
        \bottomrule 
    \end{tabular}
    }
    \caption{Impact of different types of knowledge. We investigate the impact of structural knowledge, namely the Lead images in the KB and the hierarchical relations in the KG}
    \label{fig:knowledge types}
\end{table}
Table 3 demonstrates the importance of different knowledge types in the KnowCoL-L model. The complete KnowCoL-L model achieves the best score of 31.5\% HM. Removing lead images (w/o Lead images) reduces performance to 32.5\% (seen), 27.9\% (unseen), and 30.0\% (HM), highlighting the role of visual context. Omitting hierarchical KG knowledge of `Instance of', `subclass of', and `parent taxon' further lowers scores to 32.1\% (seen), 25.3\% (unseen), and 28.3\% (HM), significantly impairing unseen entity recognition.

\subsubsection{Comparison of Fusion Functions}
\begin{table}[htbp]
    \centering
    \begin{tabular}{l c c c |c|}
        \toprule
            Fusion Method & Layer Number & Seen & Unseen & HM \\
        \midrule
            \rowcolor{gray!30} Addition & - & 34.3 & \textbf{29.1} & \textbf{31.5} \\
        \midrule
            Concatenation + MLP & 1 & 37.5 & 26.8 & 31.2\\
            Concatenation + MLP & 2 & 45.4 & 21.0 & 28.7\\
            TE & 2 & 54.4 & 16.6 & 25.5 \\
            TE + IP & 2 & \textbf{56.7} & 17.9 & 27.2\\
        \bottomrule 
    \end{tabular}

    \caption{Comparison of fusion functions for combining image and text embeddings of KnowCoL-L. Our default fusion method is addition. `TE' refers to the Transformer Encoder used in \cite{OVEN}. `TE + IP' is inspired by LLaVA~\cite{liu2023visualinstructiontuning}. In this variant, the fusion method also incorporates local image patch tokens from the ViT encoder.}
    \label{fig:fusion function}
\end{table}
As Table~\ref{fig:fusion function} shows, the default Addition method provides balanced performance, achieving the best harmonic mean 31.5\%. By using the multi-layer perceptron (MLP) method, the performance for seen entities improves up to 37.5\%, while for unseen entities decreases to 26.8\%. Using transformer encoder (TE) methods with or without local image patch embeddings significantly boost the accuracy on seen entities (54.4\% and 56.7\%), but degrade unseen entity recognition (16.6\% and 17.9\%).  The trend shows that more complex fusion methods, containing more layers or a more complex structure, lead to overfitting on seen entities and sacrificing zero-shot generalization to unseen entities. Additionally, by introducing local image patch tokens (TE + IP), performance on both seen (56.7\%) and unseen entities (17.9\%) improves compared to the TE-only method. This improvement likely occurs because local image patch tokens provide fine-grained visual details and enhance recognition ability.
\subsubsection{Comparison of KGE Methods}
\begin{table}[htbp]
    \centering
    \begin{tabular}{l l c c |c|}
        \toprule
            KGE Methods & Distance Measure & Seen & Unseen & HM \\
        \midrule
            \rowcolor{gray!30} TransE & Cosine Similarity & 34.3 & \textbf{29.1} & \textbf{31.5} \\
            TransE & Euclidean Dis. & 33.7 & 28.5 & 31.0 \\
            TransH & Cosine Similarity & \textbf{34.4} & 28.6 & 31.2 \\
            DistMult & - & 33.8 & 28.5 & 30.9\\
        \bottomrule 
    \end{tabular}
    \caption{Comparison of KGE methods for incorporating structural knowledge. Our default approach uses TransE combined with a cosine similarity metric. We compare this against alternative configurations, including TransE with Euclidean distance, TransH, and DistMult.}
    \label{fig:KGE methods}
\end{table}
Table~\ref{fig:KGE methods} compares various knowledge graph embedding methods and distance measures. Our default choice, TransE~\cite{transE} with cosine similarity, achieves the best overall performance. This setup aligns well with CLIP, which also uses cosine similarity in its contrastive learning. When replacing the cosine similarity with Euclidean distance, the HM drops from 31.5\% to 31.0\%. TransH~\cite{transH} introduces more representational flexibility by allowing entity embeddings to vary across different relations. This additional modeling capacity slightly improved the seen performance to 34.4\% but comes at the cost of lower unseen accuracy 28.6\%. The DistMult method~\cite{distMult} shows the lowest performance, suggesting it is less effective for integrating structural knowledge from the KG into the visual-textual space.

\subsubsection{Hyperparameters}
As Figure~\ref{fig:Hyperparameters} shows, we conduct experiments varying the hyperparameters $\beta_1$, $\beta_2$, latent space dimension $d_e$, and contrastive learning temperature $\tau$. Figure~\ref{fig:Hyperparameters}a shows that increasing the weight of the proxy loss $\beta_1$, aligning node embeddings with multi-modal prior knowledge, enhances model performance.  Figure~\ref{fig:Hyperparameters}b demonstrates that increasing $\beta_2$, the weight of KE loss incorporating structured prior knowledge from Wikidata, also improves performance. Additionally, we identify that the larger model (KnowCoL-H) with more prior knowledge can benefit more from the inclusion of structured prior knowledge. Figure~\ref{fig:Hyperparameters}c illustrates that larger models require higher-dimensional latent spaces to represent richer features. The temperature 0.07 performs best in our task, as shown in Figure~\ref{fig:Hyperparameters}d. 
\section{Conclusion and Future Work}
We propose KnowCoL, a novel approach to improve open-domain visual entity recognition by incorporating multi-modal prior knowledge, namely structured knowledge, textual, and visual priors. 
KnowCoL significantly enhances visual entity recognition performance, especially for zero-shot recognition of unseen entities. 
Future work includes exploring non-Euclidean spaces like hyperbolic and hyperspherical spaces to better capture the hierarchical relationships and semantic structures inherent in knowledge graphs. 

\newpage

\section{Acknowledgments}
The authors thank the International Max Planck Research School for Intelligent Systems (IMPRS-IS) for supporting Hongkuan Zhou, Yuqicheng Zhu, and Jingcheng Wu. This work was partially funded by the  European Union’s Horizon RIA research and innovation programme under grant agreement No. 101092908 (SMARTEDGE). Jingcheng Wu has been funded by the Deutsche Forschungsgemeinschaft (DFG, German Research Foundation) - SFB 1574 - Project number 471687386. The authors also gratefully acknowledge the computing time provided on the high-performance computer HoreKa by the National High-Performance Computing Center at KIT (NHR@KIT).
This center is jointly supported by the Federal Ministry of Education and Research and the Ministry of Science, Research, and the Arts of Baden-Württemberg, as part of the National High-Performance Computing (NHR) joint funding program (\url{https://www.nhr-verein.de/en/our-partners}). HoreKa is partly funded by the German Research Foundation (DFG).

\bibliography{sample-base}

\clearpage

\section{Appendix}
The following sections present detailed descriptions of the datasets and training procedures for each experimental scenario. We outline the experimental settings for each variant of our approach, including different fusion strategies for image and text embeddings, the score functions employed by various knowledge graph embedding (KGE) methods, and key hyperparameters. Additionally, we introduce our approach for extracting efficient subgraphs from Wikidata. Finally, we provide representative examples of multi-modal prior knowledge utilized in our KnowCoL approach for a clearer overview. The pseudo code for KnowCoL algorithms is listed in Algorithm~\ref{alg:KnowCoL-train} and ~\ref{alg:KnowCoL-test}. 
\section{Experiment Setting}
\subsection{Dataset}
OVEN dataset consists of 14 existing datasets, grounding their labels to Wikidata IDs. These 14 datasets are originally created for image recognition/retrieval and visual question answering. They are ImageNet21k-P~\cite{imagenet21k}, iNaturalist2017~\cite{iNaturalist}, Cars196~\cite{car196}, SUN397~\cite{SUN397}, Food101~\cite{food101}, Sports100, Aircraft~\cite{aircraft}, Oxford Flower~\cite{oxfordflower}, Google Landmarks v2~\cite{landmarks}, VQA v2~\cite{VQAv2}, Visual7W~\cite{Visual7W}, Visual Genome~\cite{VisualGenome}, OK-VQA~\cite{okvqa}, and Text-VQA~\cite{TextVQA}.

\subsection{Image and Text Query Fuser}
In this section, we elaborate on the different fusion functions in detail, namely addition, multi-layer perception (MLP), transformer encoder (TE), and transformer encoder with image patch embeddings (TE+IP). 
\subsubsection{Addition}
The default fuser we used is a simple addition operation, as it shows the best generalization ability, especially for unseen entities. The fused embedding is defined as follows:
\begin{equation}
    \mathbf{z}^\text{input} = f_{\mathbf{\theta}}(x^p) + f_{\mathbf{\lambda}}(x^t)\text{,}
    \label{equ: input_embedding_addition}
\end{equation}
where $f_{\theta}(\cdot)$ and $f_{\lambda}(\cdot)$ are the image encoder and text encoder, respectively. 
\subsubsection{Multi-layer Perception (MLP)}
The variant Multi-Layer Perceptron (MLP) fuser first concatenates the input image embedding and the text query embedding, each of dimension $d_e$, into a single embedding of dimension $2 \times d_e$. An MLP encoder then reduces this embedding dimension from $2 \times d_e$ back down to $d_e$. When the fusion layer is set to 1, the MLP consists of a single linear layer (input dimension $2 \times d_e$, output dimension $d_e$) followed by a ReLU activation function. When the fusion layer is set to 2, an additional intermediate layer of dimension $d_e$ is introduced, resulting in two linear layers, each followed by a ReLU activation. The fused embedding is represented as follows:
\begin{equation}
    \mathbf{z}^{\text{input}} = \text{MLP}\left(\text{Concat}\left(f_{\theta}(x^{p}), f_{\lambda}(x^{t})\right)\right) \text{,}
    \label{equ: input_embedding_mlp}
\end{equation}
where $\text{MLP}(\cdot)$ is the function for multi-layer perception  and $\text{Concat}(\cdot)$ is the function for concatenation. 
\subsubsection{Transformer Encoder (TE)}
The variant transformer encoder (TE) fuses the image embedding and the text embedding with a transformer encoder on top. Both embeddings are fed into the encoder, allowing them to attend to each other through self-attention. The first output embedding from the encoder is used as a global representation, capturing the fused multi-modal information.

\subsubsection{Transformer Encoder with Image Patch Embeddings (TE + IP)}
Compared to the previous approach, the local image patch embeddings from the image encoder are also fed into the transformer encoder. Similarly, the first output embedding from the encoder is used as a global representation, capturing the fused information. 

\subsection{KGE Method Settings}
Our default setting adopts the TransE method, replacing the original Euclidean distance metric with cosine similarity. The motivation for this change is that CLIP is pre-trained using cosine similarity, and adopting the same metric ensures better alignment with the embedding space learned by CLIP.  Experimental results also show that using cosine similarity improves performance. Instead of using the original margin-based formulation for negative samples, we adopt a cross-entropy-based approach, which is similar to that used in CLIP, in order to ensure that the KE loss is on the same scale as the main alignment loss. The details of other settings are given as follows.
\subsubsection{TransE + Euclidean}
In this setting, we adopt the original TransE method, using Euclidean distance and margin-based calculation for negative samples. The KE loss is defined as follows:
\begin{equation}
\begin{aligned}
\mathcal{L}_{\text{KE}}
  &= \sum_{i=1}^{N_b} \frac{1}{|\mathcal{T}_{e_i}|}
     \sum_{(h,r,t)\in\mathcal{T}_{e_i}}
     \frac{1}{|\mathcal{T}_{(h,r,t)}|}
     \sum_{(h',r,t')\in\mathcal{T}_{(h,r,t)}} \\[-2pt]
  &\quad \times\!
     \max\!\Bigl(0,\,
         \epsilon + f_s({h},{r},{t})
         - f_s({h}',{r},{t}')\Bigr)\text{,}
\end{aligned}
\label{eq:ke-loss2}
\end{equation}
where $\epsilon$ represents the margin and $f_s(\cdot)$ is the score function defined by
\begin{equation}
    f_s(h,r,t) = \|\phi(h) + \psi(r)-\phi(t)\|_2 \text{.}
\end{equation}
Here, $\phi(\cdot)$ and $\psi(\cdot)$ are functions that map nodes and relation types to corresponding embeddings. 

\subsubsection{TransH + Cosine Similarity}
TransH is an extension for TransE, which defines relation-specific hyperplane instead of direct translation in the entity space. It can be realized by altering the score function with 
\begin{equation}
\begin{split}
    f_s(h, r, t) = \| (\phi(h) - \mathbf{w}_r^\top \phi(h) \mathbf{w}_r) + \psi(r)\\
      - (\phi(t) - \mathbf{w}_r^\top \phi(t) \mathbf{w}_r) \|_2 \text{,}
\end{split}
\end{equation}
where $\mathbf{w}_r$ defines the normal vector of the hyperplane. We still use the cosine similarity and cross-entropy-based approaches to conduct contrastive learning. 
\subsubsection{DistMult}
The score function of DistMult is defined by 
\begin{equation}
\text{score}(h,r,t)
\;=\;
\phi(h)^\top \,\mathrm{diag}(\psi(r))\,\phi(t),
\end{equation}
where $\mathrm{diag}(\mathbf{r})\in\mathbb{R}^{d\times d}$ is the diagonal matrix with $\psi(r)$ on its diagonal.Similarly, we use the cosine similarity and cross-entropy-based approaches to conduct contrastive learning.

\subsection{Hyperparameters and Training Settings}
The hyperparameters used for all three variants of the KnowCoL approach, namely KnowCoL-L, KnowCoL-H, and KnowCoL-bigG, are listed as follows: 
\subsubsection{KnowCoL-L}
We adopt the OpenCLIP ViT-L/14 image encoder as our backbone and train for 15 epochs with a batch size of 4096. Optimization is performed using AdamW (learning rate 0.0001, $\beta_1=1.0$, $\beta_2=1.0$, weight decay $1 \times 10^{-4}$) together with a cosine-annealing learning-rate schedule. We set the embedding dimension $d_e$ to 768 and the temperature of contrastive learning $\tau$ to 0.07. In our knowledge-graph module, each entity is associated with up to 50 triplets, and we randomly sample 25 negative triplets per entity. The model was trained using 4 NVIDIA H100 GPUs, each with 80GB of memory, over a total duration of approximately 22 hours. The training was performed using mixed precision (FP16).

\subsubsection{KnowCoL-H}
We adopt the OpenCLIP ViT-H/14 image encoder as our backbone. The model is trained for 20 epochs with a batch size of $4096$. The AdamW optimizer is employed with a learning rate of $1\times10^{-3}$, cosine annealing scheduling, and weight decay of $1\times10^{-4}$. The hyperparameters $\beta_1$, $\beta_2$, latent dimension $d_e$, and contrastive learning temperature $\tau$ are set as $1.0$, $1.0$, $1024$, and $0.07$, respectively. For the knowledge graph, each entity is associated with at most $50$ triplets, and we draw $25$ negative samples per entity. The model is trained with 8 NVIDIA H100 GPUs with 80GB memory. The total training time is around 25 hours. The training was performed using mixed precision (FP16).

\subsubsection{KnowCoL-bigG}
We adopt the OpenCLIP ViT-bigG/14 image encoder as our backbone. The model is trained for 25 epochs with a batch size of $4096$. The AdamW optimizer is employed with a learning rate of $1\times10^{-3}$, cosine annealing scheduling, and weight decay of $1\times10^{-4}$. The hyperparameters $\beta_1$, $\beta_2$, latent dimension $d_e$, and contrastive learning temperature $\tau$ are set as $1.0$, $1.0$, $1024$, and $0.07$, respectively. For the knowledge graph, each entity is associated with at most $50$ triplets, and we draw $25$ negative samples per entity. The model is trained with 8 Nvidia H200 GPUs with 140GB of memory. The total training time is about 30 hours. The training was performed using mixed precision (FP16).

\section{Structure Knowledge Extraction}
Wikidata comprises billions of entities and thousands of properties, making it computationally infeasible to utilize the entire knowledge graph for model training. Moreover, the noise within Wikidata poses an additional hurdle, potentially obstructing the learning process and affecting model performance. To mitigate this challenge, we extract a subgraph from Wikidata that encapsulates the essential information required for the OVEN benchmark. We mainly focus on the entity-to-entity relationships among all entities occurring in the OVEN benchmark, plus the hierarchical relations `subclass of' (P279), `instance of' (P31), and `parent taxon'(P171) to preserve the structural integrity of the knowledge representation. 

\label{sec:Structure Knowledge Extraction}
\begin{algorithm}[ht]
\caption{Subgraph Extraction from Wikidata}
\label{alg:subgraph_extraction}
\begin{algorithmic}[1]
\STATE Given:
\begin{itemize}
    \item Original Wikidata KG $\mathcal{G}=(\mathcal{E}, \mathcal{R}, \mathcal{T})$
    \item Dataset entity set $\mathcal{E}_\text{data}$
    \item Hierarchical relations $p_1$ (`subclass of'), $p_2$ (`instance of'), and optionally $p_3$ (`parent taxon').
\end{itemize}
\STATE Initialize $\mathcal{E}_\text{data}^+ \leftarrow \mathcal{E}_\text{data}$
\FOR{each entity $e \in \mathcal{E}_\text{data}$}
    \STATE Find superclasses: $S_e = \{ e' \mid (e, p_1, e') \in \mathcal{T} \lor (e, p_2, e') \in \mathcal{T} \}$
    \STATE Update $\mathcal{E}_\text{data}^+ \leftarrow \mathcal{E}_\text{data}^+ \cup S_e$
\ENDFOR
\STATE Initialize $\mathcal{T}_\text{sub} \leftarrow \emptyset$, $\mathcal{R}_\text{sub} \leftarrow \emptyset$
\FOR{each triple $(e_1, r, e_2) \in \mathcal{T}$}
    \IF{$e_1 \in \mathcal{E}_\text{data}^+$ and $e_2 \in \mathcal{E}_\text{data}^+$}
        \STATE $\mathcal{T}_\text{sub} \leftarrow \mathcal{T}_\text{sub} \cup \{(e_1, r, e_2)\}$
        \STATE $\mathcal{R}_\text{sub} \leftarrow \mathcal{R}_\text{sub} \cup \{r\}$
    \ENDIF
\ENDFOR
\STATE Define induced subgraph: $\mathcal{G}_\text{sub} = (\mathcal{E}_\text{data}^+, \mathcal{R}_\text{sub}, \mathcal{T}_\text{sub})$
\RETURN $\mathcal{G}_\text{sub}$
\end{algorithmic}
\end{algorithm}

The Wikidata knowledge graph is denoted as $\mathcal{G} = (\mathcal{E}, \mathcal{R}, \mathcal{T})$, where $\mathcal{E}$, $\mathcal{R}$ represents the entity set and predicates set, respectively. $\mathcal{T} \subseteq \mathcal{E} \times \mathcal{R} \times \mathcal{E}$ is the set of triples ($e_1$, $r$, $e_2$) where $e_1$, $e_2 \in \mathcal{E}$ and $r \in \mathcal{R}$. 

The entity set of the dataset is denoted as $\mathcal{E}$. We first expand it by incorporating all superclasses of the entities in ${E}$, as defined by the `subclass of' $p_1$ and the `instance of' $p_2$ relation in Wikidata. The resulting expanded set, denoted as $E^+$, is defined as:
\begin{equation}
    \mathcal{E}_\text{data}^+ = {\mathcal{E}_\text{data}} \cup \{ e^{\prime} \mid e \in \mathcal{E}_\text{data }, (e, p_1, e^{\prime}) \in \mathcal{G} \lor (e, p_2, e^{\prime}) \in \mathcal{G}\}\text{,}
\end{equation}
where $e^{\prime}$ denotes a superclass of entity $e$. 

We define the subgraph $\mathcal{G}_\text{sub}$ as the induced subgraph containing all entities in $\mathcal{E}^+_\text{data}$ and all relations that exist between these entities. Formally, the subgraph $\mathcal{G}_\text{sub}$ is defined as:
\begin{equation}
    \mathcal{G}_\text{sub} =(\mathcal{E}^+_\text{data}\mathcal{R}_\text{sub}, \mathcal{T}_\text{sub})
\end{equation}
where 
\begin{equation}
\begin{aligned}
\mathcal{R}_\text{sub} &= \{r \mid \exists e_1, e_2 \in \mathcal{E}^+_\text{data}, (e_1, r, e_2) \in \mathcal{T} \}, \\
\mathcal{T}_\text{sub} &= \{ (e_1, r, e_2) \in \mathcal{T} \mid e_1, e_2 \in \mathcal{E}^+_\text{data} \land r \in \mathcal{R}\}.
\end{aligned}
\end{equation}
This subgraph $\mathcal{G}_\text{sub}$ contains the relations between entities existing in the OVEN benchmark and the hierarchical relations between the entities and their super classes. The pseudo code can be seen in Algorithm~\ref{alg:subgraph_extraction}.

\begin{algorithm*}[ht]
\caption{Knowledge-Guided Contrastive Learning - Training}
\label{alg:KnowCoL-train}
\begin{algorithmic}[1]
\STATE Given:
\begin{itemize}
    \item Dataset $D=\{(x^p_i, x^t_i), e_i\}_{i=1}^{N}$
    \item Knowledge Base $\mathcal{K} = \{(t_{e_i},I_{e_i})\}_{i=1}^{|\mathcal{E}|}$, where $t_{e_i}$ is the text description and $I_{e_i}$ is the lead images set of entity $e_i$
    \item Knowledge Graph $\mathcal{G} = (\mathcal{E}, \mathcal{R}, \mathcal{T})$
    \item The model $\mathcal{F}= \{f_\theta, f_\lambda, f_\gamma\, \phi, \psi\}$ with pre-trained image and text encoders $f_\theta$,  $f_\lambda$, fusion function $f_\gamma$, and entity and relation mapping function $\phi,\psi$
\end{itemize}
\STATE Initialize projection layers of encoders $f_\theta$, $f_\lambda$, and fusion $f_\gamma$, the other parameters of the encoders remain fixed.
\STATE Randomly initialize entity embeddings lookup table $\{ \phi(e_i)\}_{i=1}^{|\mathcal{E}|} $ and relation embeddings lookup table $\{\psi(r_i)\}_{i=1}^{|\mathcal{R}|}$
\WHILE{not converged}
    \STATE Sample batch $\{(x^p_i, x^t_i, e_i)\}_{i=1}^{N_b}$ from $D$
    \STATE $\mathbf{z}_i^\text{input} = f_\gamma(f_\theta(x_i^p), f_\lambda(x_i^t))$
    \STATE $\mathbf{z}_i^\text{entityText} = f_\lambda(t_{e_i})$
    \STATE $\mathbf{z}_i^\text{entityImage} = \frac{1}{|I_e|}\sum_{a \in I_e} f_\theta(a)$ if  $I_{e_i}\ne \emptyset$ else $z_i^\text{entityText}$ 
    \STATE Compute alignment loss $L_a$, proxy loss $L_p$, and knowledge graph embedding loss $L_{KE}$ 
    \STATE Compute total loss $L = L_a + \beta_1 L_p + \beta_2 L_{KE}$
    \STATE Update parameters $\theta,\lambda, \gamma$ based on $L$
\ENDWHILE
\end{algorithmic}
\end{algorithm*}

\begin{algorithm*}[ht]
\caption{Knowledge-Guided Contrastive Learning - Inference}
\label{alg:KnowCoL-test}
\begin{algorithmic}[1]
\STATE Given
\begin{itemize}
    \item Image $x^p$
    \item Text query $x^t$ 
\end{itemize}
\STATE Compute $\mathbf{z}^\text{input} = f_\gamma(f_\theta(x^p), f_\lambda(x^t))$
\STATE Return predicted entity $e^* = \arg\max_{e_j \in \mathcal{E}} \text{sim}\left(\mathbf{z}^{\text{input}},\,\frac{1}{2} \left( \mathbf{z}^{\text{entityText}}_j + \mathbf{z}^{\text{entityImage}}_j\right) \right) $
\end{algorithmic}
\end{algorithm*}

\section{Additional Experiments}
The OVEN dataset consists of two splits: the entity split, derived from image recognition datasets, and the query split, originating from visual question answering datasets. In this paper, we primarily focus on the entity split. For completeness, we also show results on the query split below. We observed that the OVEN training data includes a limited amount of query split data, making this split significantly underrepresented. In the training dataset, 4,926,314 samples belong to the entity split while only 32,255 samples belong to the query split. To address this imbalance, we duplicate the query split data in each training epoch by 50$\times$, which substantially improves performance on the query-based test set. We also observe that generative models such as BLIP-v2 and PaLI perform better on the query split than on the entity split, despite their inability to consistently predict the correct entity. This may be because these models are trained on visual question answering (VQA) tasks and are therefore more aligned with the query split format. Alternatively, the improved performance might be partially attributed to data leakage or overlap with the VQA-style training data.

\begin{table}[ht]
    \centering
    \begin{tabular}{c| c c c c}
    \toprule
        \multirow{2}{*}{Methods} & \multirow{2}{*}{\makecell{Duplicate \\ Numbers}} &\multicolumn{3}{c}{Query Split}\\
         &   & Seen & Unseen & HM \\
    \midrule
        KnowCoL-L & 1 & 8.5 & 6.1 & 7.1\\
        KnowCoL-L & 10 & 11.7 & 7.3 & 8.9\\
        KnowCoL-L & 50 & 16.2 & 8.0 & 10.7\\
        \midrule
        KnowCoL-H & 50 &18.5 & 10.5 & 13.4\\
        KnowCoL-bigG & 50 & 21.7 & 12.1 & 15.6\\
    \bottomrule 
    \end{tabular}
    \caption{Query Split Results}
    \label{tab:placeholder}
\end{table}

\section{Multi-modal Prior Knowledge of \\ Sample Entities}
Table~\ref{tab:wikidata_kg} presents representative examples of prior knowledge used in the proposed model. Specifically, we provide three illustrative examples each for animals, cars, landmarks, and people. For each example, we include a lead image, a concise textual description, and associated structural knowledge from Wikidata. The structured relationships relevant to these entities, such as instance of' (P31), subclass of' (P279), and `parent taxon' (P171), among other significant properties, are detailed in the table. This structured knowledge helps enrich entity representation, facilitating semantic disambiguation and enhancing the generalization capabilities of our KnowCoL framework.

\begin{table*}[htbp]
\centering
\caption{Prior Knowledge of Sample Entities: for each entity we show its lead image, a brief textual description, and its structural knowledge as key Wikidata triples (with that entity as \textbf{head}). }
\resizebox{\linewidth}{!}{%
\begin{tabular}{l c p{7cm} | c c}
\toprule
\multicolumn{3}{c|}{Entity} & Predicate & Entity \\
\cline{1-3}
Name (QID) & Lead Image &  Text Description & Name (PID) & Name (QID)\\
\midrule

\multirow{5}{*}{\makecell[l]{Mountain Hare \\ (\textcolor{light-orange}{Q180035})}} & \multirow{5}{*}{\includegraphics[width=2cm]{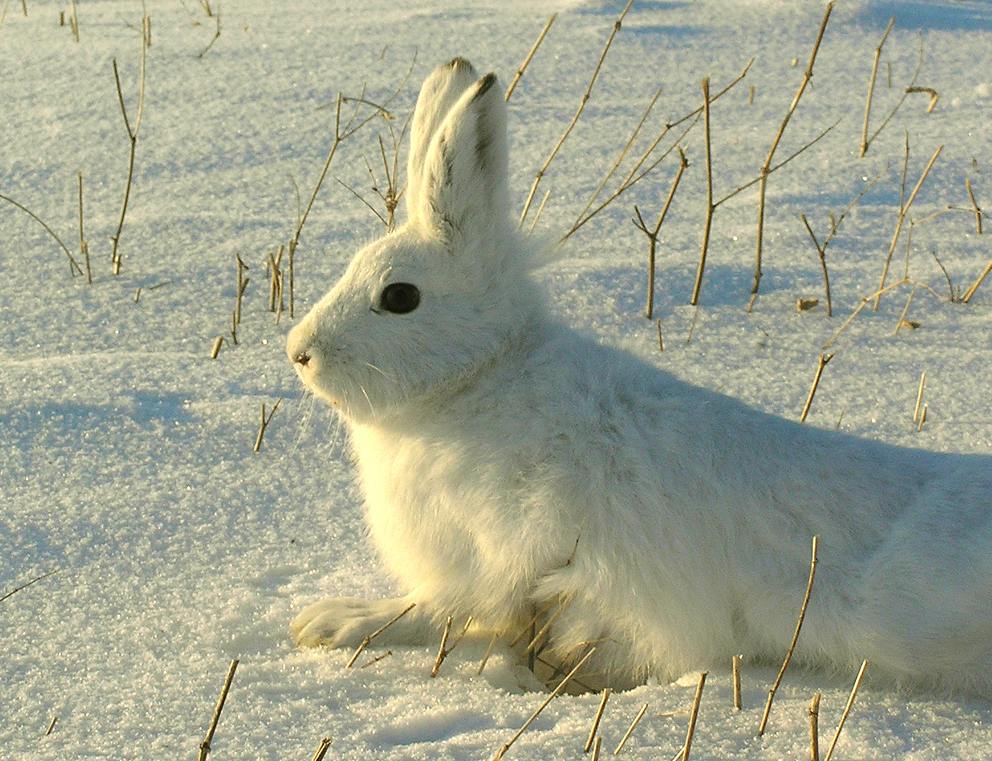}} & \multirow{5}{*}{%
    \parbox{7cm}{
        The mountain hare (Lepus timidus), also known as blue hare, tundra hare, variable hare, white hare, snow hare, alpine hare, and Irish hare, is a species of Palearctic hare that is largely adapted to polar and mountainous habitats...
    }
} & Instance of (\textcolor{light-purple}{P31}) & Taxon (\textcolor{light-orange}{Q16521}) \\
&   &  & Taxon Rank (\textcolor{light-purple}{P105}) & Species (\textcolor{light-orange}{Q7432}) \\
&  &  & Parent Taxon (\textcolor{light-purple}{P171}) & Lepus (\textcolor{light-orange}{Q3830767}) \\
&   &  & Diel Cycle (\textcolor{light-purple}{P9566}) & Nocturnal (\textcolor{light-orange}{Q101029366}) \\
&  &  & Habitat (\textcolor{light-purple}{P2974}) & Grassland (\textcolor{light-orange}{Q1006733}) \\

\midrule

\multirow{5}{*}{\makecell[l]{Platypus \\ (\textcolor{light-orange}{Q15343})}} & \multirow{5}{*}{\includegraphics[width=2cm]{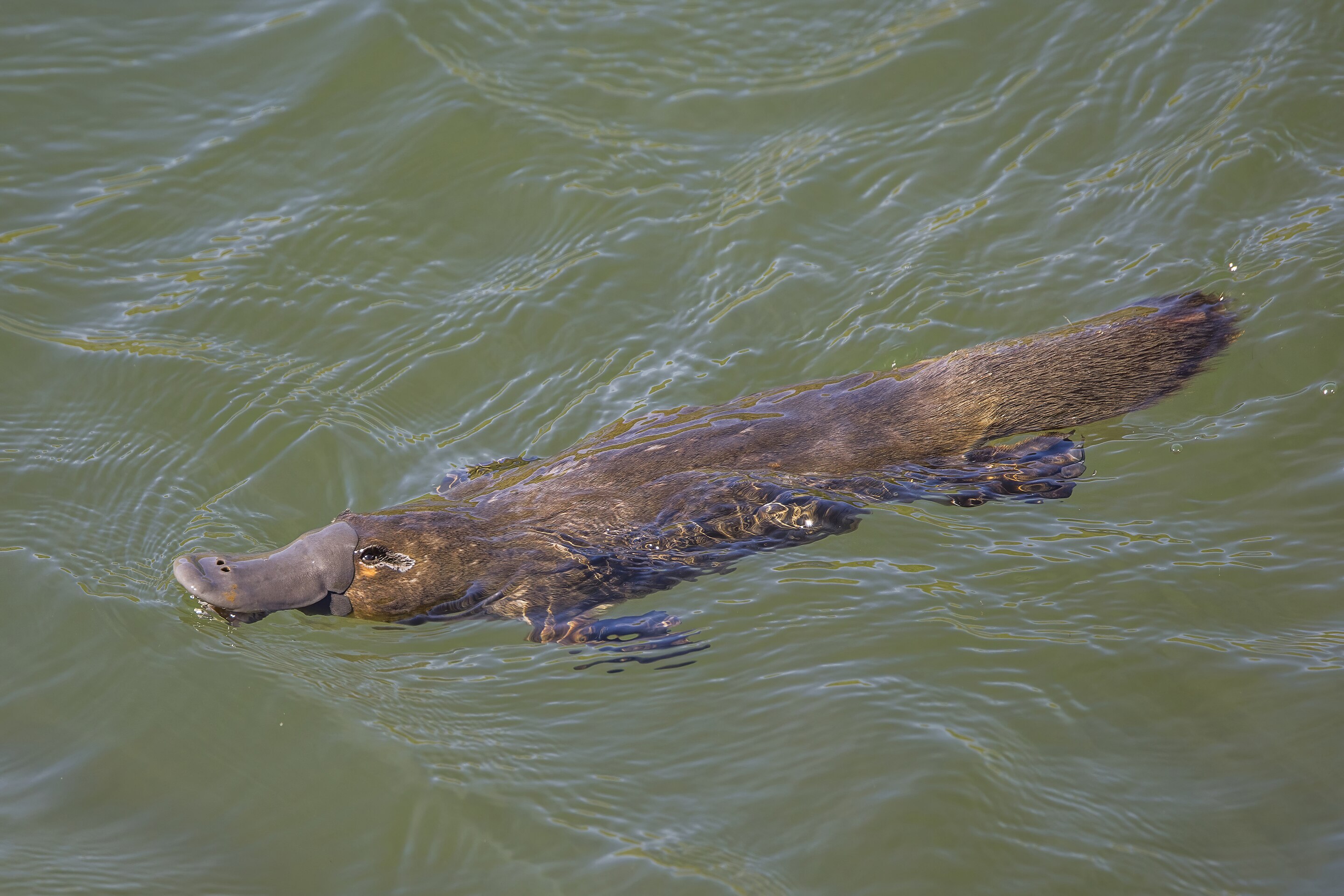}} & \multirow{5}{*}{%
    \parbox{7cm}{
        The platypus (Ornithorhynchus anatinus), sometimes referred to as the duck-billed platypus, is a semiaquatic, egg-laying mammal endemic to eastern Australia, including Tasmania. The platypus is the sole living representative or ...
    }
} & Instance of (\textcolor{light-purple}{P31}) & Taxon (\textcolor{light-orange}{Q16521}) \\
&   &  & Taxon Rank (\textcolor{light-purple}{P105}) & Species (\textcolor{light-orange}{Q7432}) \\
&  &  & Parent Taxon (\textcolor{light-purple}{P171}) & Ornithorhyn. (\textcolor{light-orange}{Q3745848}) \\
&   &  & Endemic To (\textcolor{light-purple}{P183}) & Australia (\textcolor{light-orange}{Q408}) \\
&  &  & Characteristic (\textcolor{light-purple}{P1552}) & Oviparity (\textcolor{light-orange}{Q212306}) \\

\midrule

\multirow{5}{*}{\makecell[l]{Morpho \\ Menelaus \\ (\textcolor{light-orange}{Q545709})}} & \multirow{5}{*}{\includegraphics[width=2cm]{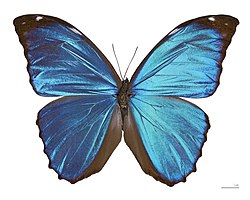}} & \multirow{5}{*}{%
    \parbox{7cm}{
        The Menelaus blue morpho (Morpho menelaus) is one of thirty species of butterfly in the subfamily Morphinae. Its wingspan is approximately 12 cm (4.7''), and its dorsal forewings and hindwings are a bright, iridescent blue edged with black ...
    }
} & Instance of (\textcolor{light-purple}{P31}) & Taxon (\textcolor{light-orange}{Q16521}) \\
&   &  & Taxon Rank (\textcolor{light-purple}{P105}) & Species (\textcolor{light-orange}{Q7432}) \\
&  &  & Parent Taxon (\textcolor{light-purple}{P171}) & Morpho (\textcolor{light-orange}{Q645804}) \\
&   &  & Color (\textcolor{light-purple}{P462}) & Blue (\textcolor{light-orange}{Q1088}) \\
&  &  & Has Part(s) (\textcolor{light-purple}{P527}) & Insect Wing  (\textcolor{light-orange}{Q276572}) \\

\midrule

\multirow{5}{*}{\makecell[l]{BMW X5 \\ (\textcolor{light-orange}{Q796778})}} & \multirow{5}{*}{\includegraphics[width=2cm]{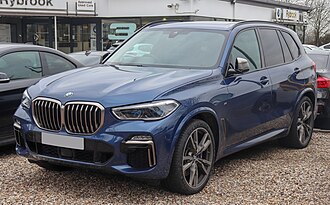}} & \multirow{5}{*}{%
    \parbox{7cm}{
        The BMW X5 is a mid-size luxury crossover SUV produced by BMW. The X5 made its debut in 1999 as the E53 model. It was BMW's first SUV. At launch, it featured all-wheel drive and was available with either a manual or ...
    }
} & Instance of (\textcolor{light-purple}{P31}) & Automobile M. (\textcolor{light-orange}{Q3231690}) \\
&   &  & Manufacturer (\textcolor{light-purple}{P176}) & BMW (\textcolor{light-orange}{Q26678}) \\
&  &  & Subclass of (\textcolor{light-purple}{P279}) & Executive Car (\textcolor{light-orange}{Q1357619}) \\
&   &  & Subclass of (\textcolor{light-purple}{P279}) & Car (\textcolor{light-orange}{Q1420})\\
&  &  & Subclass of (\textcolor{light-purple}{P279}) & Road Vehicle (\textcolor{light-orange}{Q1515493}) \\

\midrule

\multirow{5}{*}{\makecell[l]{Tesla Model S \\ (\textcolor{light-orange}{Q1463050})}} & \multirow{5}{*}{\includegraphics[width=2cm]{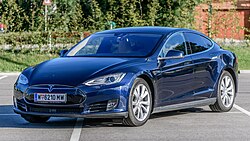}} & \multirow{5}{*}{%
    \parbox{7cm}{
        The Tesla Model S is a battery-electric, four-door full-size car produced by the American automaker Tesla since 2012. The automaker's second vehicle and longest-produced model, the Model S has both received mixed reviews from critics ...
    }
} & Instance of (\textcolor{light-purple}{P31}) & Automobile M. (\textcolor{light-orange}{Q3231690}) \\
&   &  & Manufacturer (\textcolor{light-purple}{P176}) & Tesla, Inc. (\textcolor{light-orange}{Q478214}) \\
&  &  & Follows (\textcolor{light-purple}{P155}) & Tesla Roadster (\textcolor{light-orange}{Q210893}) \\
&   &  & Followed by (\textcolor{light-purple}{P156}) & Model X (\textcolor{light-orange}{Q1634161})\\
&  &  & Subclass of (\textcolor{light-purple}{P279}) & Electric Car (\textcolor{light-orange}{Q193692}) \\

\midrule

\multirow{5}{*}{\makecell[l]{Volkswagen \\ Beetle \\ (\textcolor{light-orange}{Q1968742})}} & \multirow{5}{*}{\includegraphics[width=2cm]{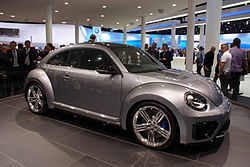}} & \multirow{5}{*}{%
    \parbox{7cm}{
        The Volkswagen Beetle, officially the Volkswagen Type 1, is a small family car produced by the German company Volkswagen from 1938 to 2003. One of the most iconic cars in automotive history, the Beetle is noted for its distinctive shape...
    }
} & Instance of (\textcolor{light-purple}{P31}) & Automobile M. (\textcolor{light-orange}{Q3231690}) \\
&  &  & Instance of (\textcolor{light-purple}{P31}) & Vehicle M. (\textcolor{light-orange}{Q29048322}) \\
&   &  & Brand (\textcolor{light-purple}{P1716}) & Volkswagen(\textcolor{light-orange}{Q246}) \\
&  &  & Origin Country (\textcolor{light-purple}{P495}) & Germany (\textcolor{light-orange}{Q183}) \\
&   &  & Powered by (\textcolor{light-purple}{P516}) & Gas. Engine (\textcolor{light-orange}{Q502048})\\

\midrule

\multirow{5}{*}{\makecell[l]{Taj \\ Mahal \\ (\textcolor{light-orange}{Q9141})}} & \multirow{5}{*}{\includegraphics[width=2cm]{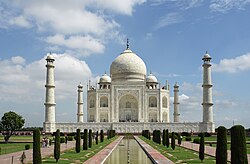}} & \multirow{5}{*}{%
    \parbox{7cm}{
        The Taj Mahal is an ivory-white marble mausoleum on the right bank of the river Yamuna in Agra, Uttar Pradesh, India. It was commissioned in 1631 by the fifth Mughal emperor, Shah Jahan to house the tomb of his beloved wife ...
    }
} & Instance of (\textcolor{light-purple}{P31}) & Tomb (\textcolor{light-orange}{Q381885}) \\
&  &  & Instance of (\textcolor{light-purple}{P31}) & Tourist Attr.(\textcolor{light-orange}{Q570116}) \\
&   &  & Religion (\textcolor{light-purple}{P140}) & Islam(\textcolor{light-orange}{Q432}) \\
&  &  & Country (\textcolor{light-purple}{P17}) & India (\textcolor{light-orange}{Q668}) \\
&   &  & Architect (\textcolor{light-purple}{P84}) & Ahmad Lahori (\textcolor{light-orange}{Q2551253})\\

\midrule

\multirow{5}{*}{\makecell[l]{Potala \\ Palace \\ (\textcolor{light-orange}{Q71229})}} & \multirow{5}{*}{\includegraphics[width=2cm]{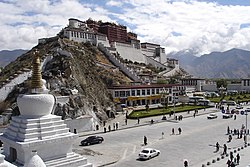}} & \multirow{5}{*}{%
    \parbox{7cm}{
        Potala Palace is the name of a museum in the Tibet Autonomous Region of China previously a palace of the Bö sovereign in Ü, the Dalai Lama, in the dzong-style, in Lhasa, capital of Bod and historically Ü. It was the winter palace of the ... 
    }
} & Instance of (\textcolor{light-purple}{P31}) & Palace (\textcolor{light-orange}{Q16560}) \\
&  &  & Instance of (\textcolor{light-purple}{P31}) & Tourist Attr.(\textcolor{light-orange}{Q570116}) \\
&   &  & Has Use (\textcolor{light-purple}{Q71229}) & Official Resi. (\textcolor{light-orange}{Q481289}) \\
&  &  & Different From (\textcolor{light-purple}{P1889}) & State Housing (\textcolor{light-orange}{Q7603702}) \\
&   &  & Country (\textcolor{light-purple}{P17}) & China (\textcolor{light-orange}{Q148})\\

\midrule

\multirow{5}{*}{\makecell[l]{Faroe \\ Islands \\ (\textcolor{light-orange}{Q4628})}} & \multirow{5}{*}{\includegraphics[width=2cm]{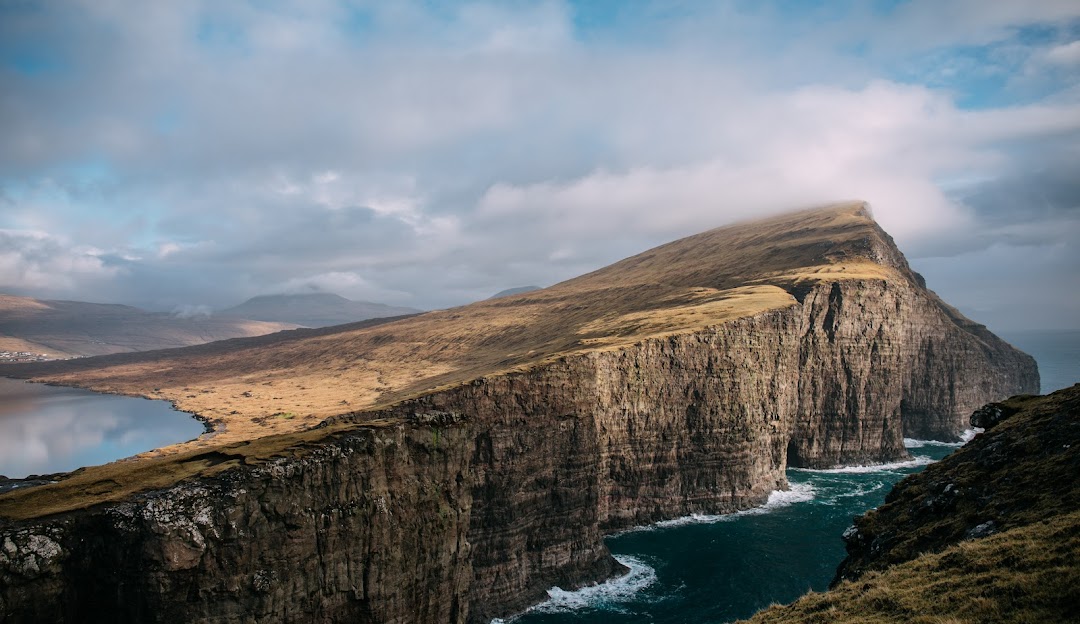}} & \multirow{5}{*}{%
    \parbox{7cm}{
        The Faroe Islands are an archipelago in the North Atlantic Ocean and an autonomous territory of the Kingdom of Denmark. Located between Iceland, Norway, and the United Kingdom, the islands have a population of 54,900 as ...
    }
} & Instance of (\textcolor{light-purple}{P31}) & Country (\textcolor{light-orange}{Q6256}) \\
&  &  & Country (\textcolor{light-purple}{P17}) & Denmark (\textcolor{light-orange}{Q35}) \\
&   &  & Part of (\textcolor{light-purple}{P361}) & Northern Europe  (\textcolor{light-orange}{Q27479}) \\
&  &  & Continent (\textcolor{light-purple}{P30}) & Europe (\textcolor{light-orange}{Q46}) \\
&   &  & Capital (\textcolor{light-purple}{P36}) & Tórshavn (\textcolor{light-orange}{Q10704})\\

\midrule

\multirow{5}{*}{\makecell[l]{Stephen \\ Curry \\ (\textcolor{light-orange}{Q352159})}} & \multirow{5}{*}{\includegraphics[width=1.45cm]{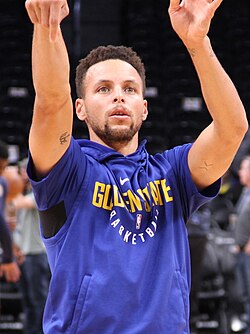}} & \multirow{5}{*}{%
    \parbox{7cm}{
        Wardell Stephen Curry II, born March 14, 1988, also known as Steph Curry, is an American professional basketball player and point guard for the Golden State Warriors of the National Basketball Association (NBA).
    }
} & Instance of (\textcolor{light-purple}{P31}) & Human (\textcolor{light-orange}{Q5}) \\
&  &  & Sex/Gender (\textcolor{light-purple}{P21}) & Male (\textcolor{light-orange}{Q6581097}) \\
&   &  & Citizenship (\textcolor{light-purple}{P27}) & United States (\textcolor{light-orange}{Q30}) \\
&  &  & Occupation (\textcolor{light-purple}{P106}) & Basketball P. (\textcolor{light-orange}{Q3665646}) \\
&   &  & Father (\textcolor{light-purple}{P22}) & Dell Curry (\textcolor{light-orange}{Q1184381})\\

\midrule

\multirow{5}{*}{\makecell[l]{Lady \\ Gaga \\ (\textcolor{light-orange}{Q19848})}} & \multirow{5}{*}{\includegraphics[width=1.45cm]{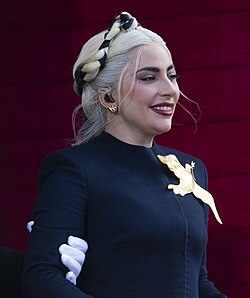}} & \multirow{5}{*}{%
    \parbox{7cm}{
        Stefani Joanne Angelina Germanotta[a] (born March 28, 1986), known professionally as Lady Gaga, is an American singer, songwriter, and actress. She is an influential figure in popular music, known for her image reinventions and ... 
    }
} & Instance of (\textcolor{light-purple}{P31}) & Human (\textcolor{light-orange}{Q5}) \\
&  &  & Sex/Gender (\textcolor{light-purple}{P21}) & Female (\textcolor{light-orange}{Q6581072}) \\
&   &  & Citizenship (\textcolor{light-purple}{P27}) & United States (\textcolor{light-orange}{Q30}) \\
&  &  & Occupation (\textcolor{light-purple}{P106}) & Singer (\textcolor{light-orange}{Q177220}) \\
&   &  & Occupation (\textcolor{light-purple}{P106}) & Songwriter (\textcolor{light-orange}{Q753110})\\

\midrule

\multirow{5}{*}{\makecell[l]{Milan \\ Kundera \\ (\textcolor{light-orange}{Q93166})}} & \multirow{5}{*}{\includegraphics[width=1.75cm]{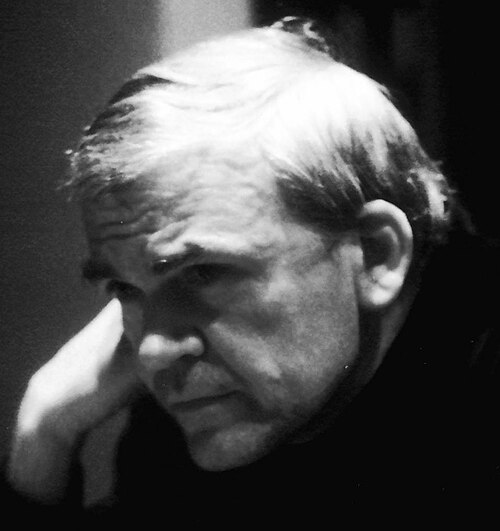}} & \multirow{5}{*}{%
    \parbox{7cm}{
        Milan Kundera was a Czech and French novelist. Kundera went into exile in France in 1975, acquiring citizenship in 1981. His Czechoslovak citizenship was revoked in 1979, but he was granted Czech citizenship in 2019...
    }
} & Instance of (\textcolor{light-purple}{P31}) & Human (\textcolor{light-orange}{Q5}) \\
&  &  & Sex/Gender (\textcolor{light-purple}{P21}) & Male (\textcolor{light-orange}{Q6581097}) \\
&   &  & Citizenship (\textcolor{light-purple}{P27}) & France (\textcolor{light-orange}{Q142}) \\
&  &  & Occupation (\textcolor{light-purple}{P106}) & Writer (\textcolor{light-orange}{Q36180}) \\
&   &  & Instrument (\textcolor{light-purple}{P1303}) & Piano (\textcolor{light-orange}{Q5994})\\

\bottomrule
\end{tabular}
}
\label{tab:wikidata_kg}
\end{table*}




\subsection{Qualitative Analysis}
\setlength{\tabcolsep}{1mm}
\begin{table*}[!t]
    \centering
    \begin{tabular}{c c c c c c}
        \toprule
        \multirow{2}{*}{Input Image} & \multirow{2}{*}{Text Query} & \multicolumn{3}{c}{Prediction} & \multirow{2}{*}{Ground Truth}\\
        & & 1 & 2 & 3 & \\ 
        \midrule
        {\makecell{\includegraphics[height=1.5cm]{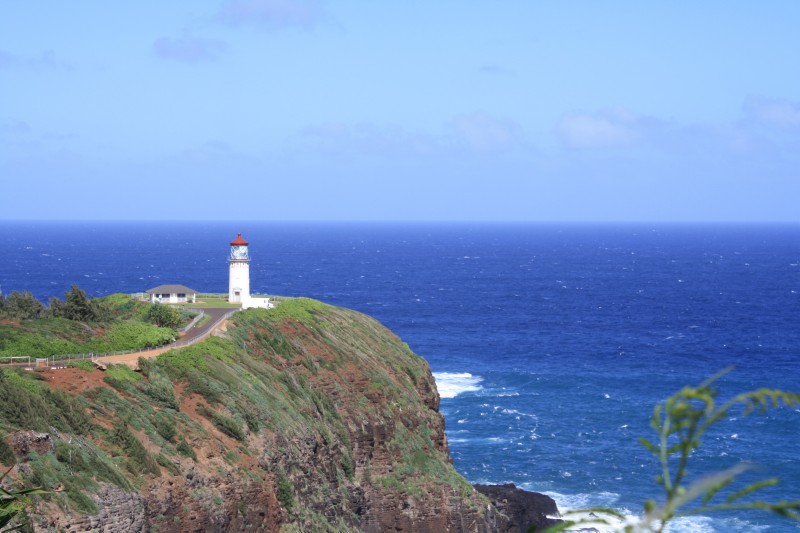}}}& \makecell{what is this building called?}& \makecell{Q6406516\\ \imgbordergreen{\includegraphics[height=1.5cm]{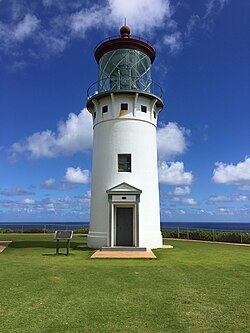}}} & \makecell{Q15278727\\ \imgborderred{\includegraphics[height=1.5cm]{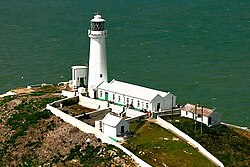}}} & \makecell{Q7637054 \\ \imgborderred{\includegraphics[height=1.5cm]{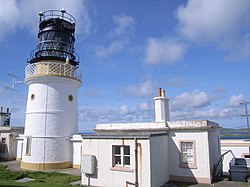}}} & \makecell{Q6406516\\ \includegraphics[height=1.5cm]{images/lead_images/Q6406516.jpg}}\\

        \midrule
        
        \makecell{\includegraphics[height=2cm]{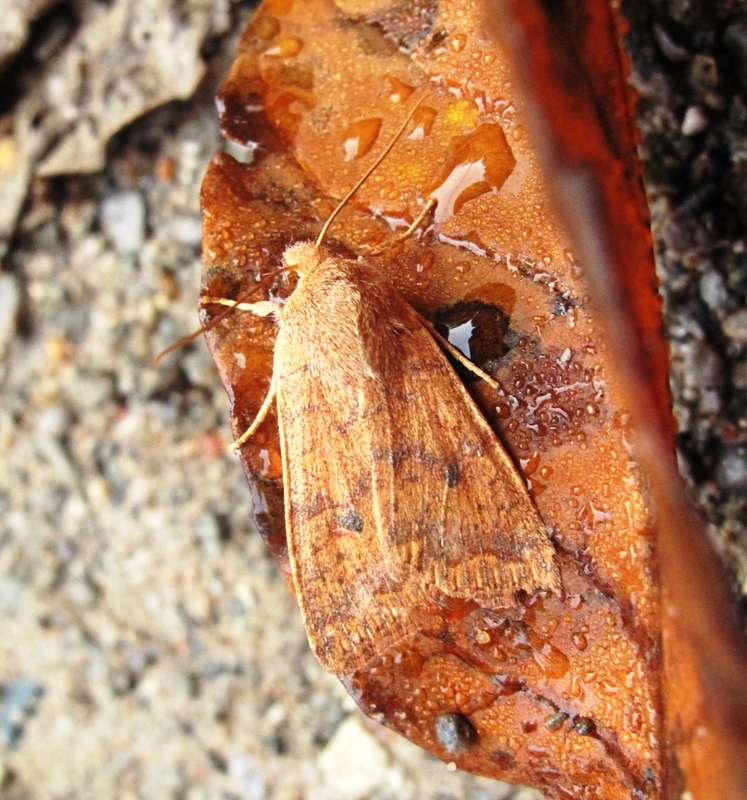}}& \makecell{which species of insect is this?}& \makecell{Q5135458\\ \imgborderred{\includegraphics[height=1.5cm]{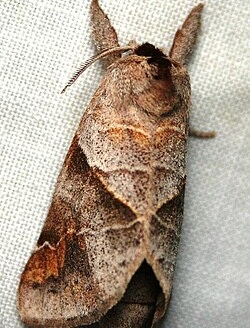}}} & \makecell{Q5379231 \\ \imgborderred{\includegraphics[height=1.5cm]{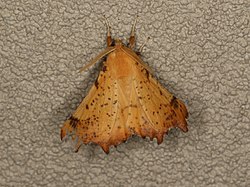}}} & \makecell{Q4694221 \\ \imgbordergreen{\includegraphics[height=1.5cm]{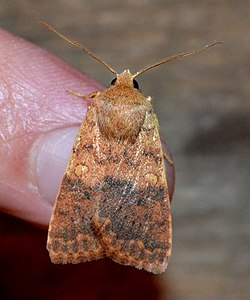}}} & \makecell{Q4694221 \\ \includegraphics[height=1.5cm]{images/lead_images/Q4694221.jpg}}\\
        \midrule
        
        \makecell{\includegraphics[height=1.5cm]{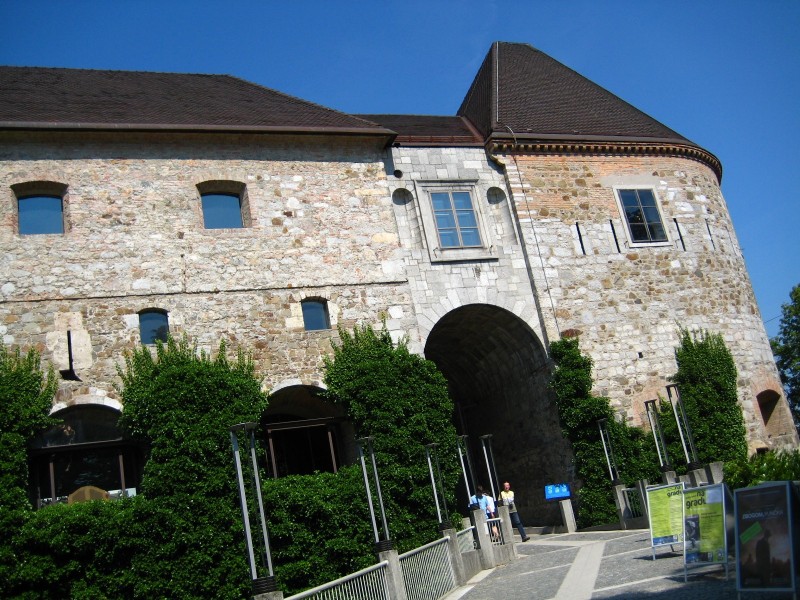}}& \makecell{where is this place?}& \makecell{Q1014274 \\ \imgborderred{\includegraphics[height=1.5cm]{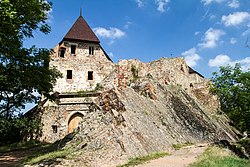}}} & \makecell{Q2193788 \\ \imgborderred{\includegraphics[height=1.5cm]{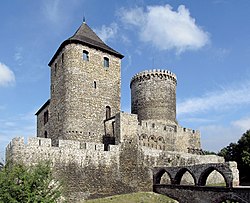}}} & \makecell{Q1012308 \\ \imgborderred{\includegraphics[height=1.5cm]{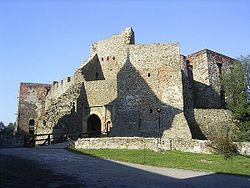}}} & \makecell{Q2075156 \\ \includegraphics[height=1.5cm]{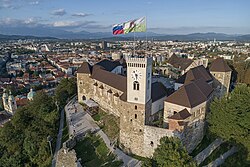}}\\

        \midrule
        
        \makecell{\includegraphics[height=1.5cm]{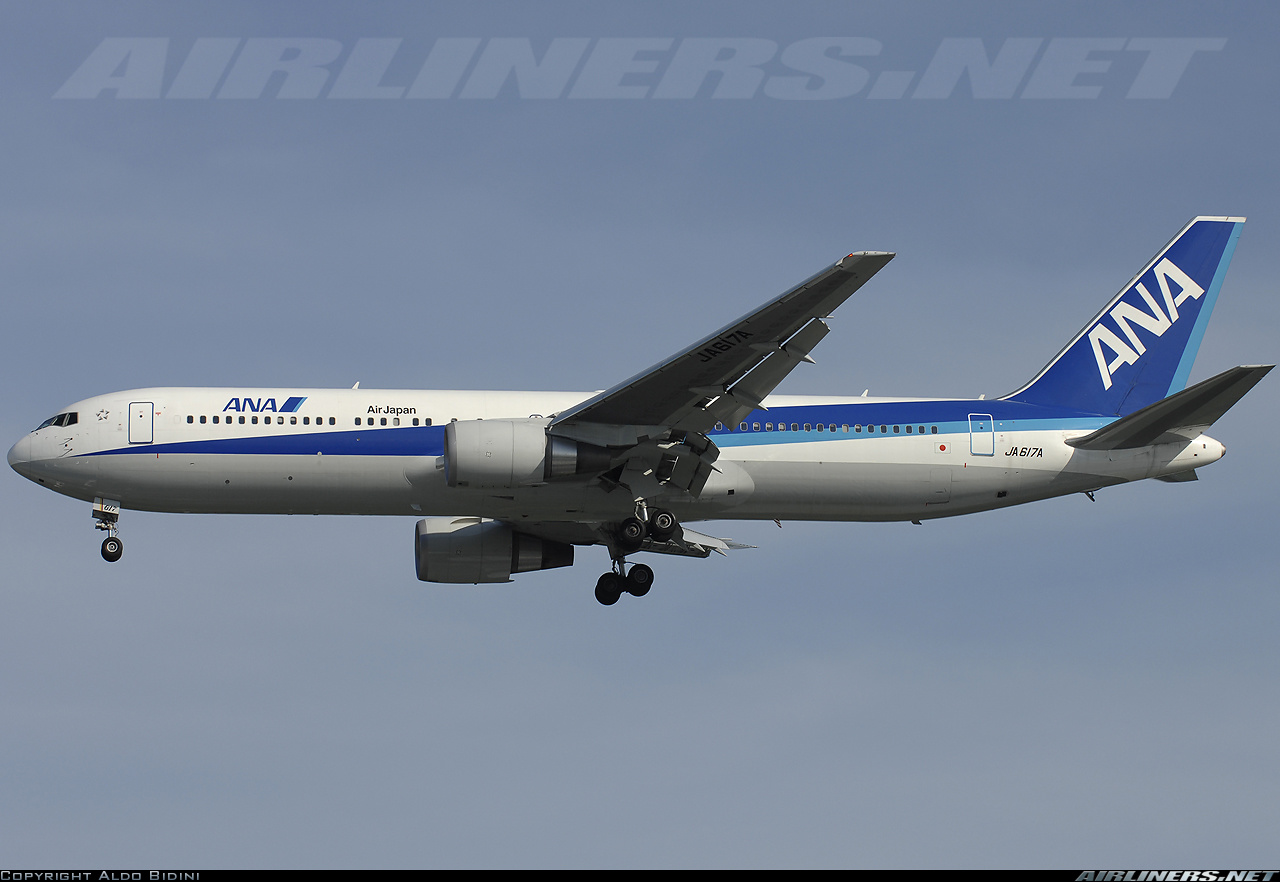}}& \makecell{what is the model of this aircraft?}& \makecell{Q6423  \\ \imgbordergreen{\includegraphics[height=1.5cm]{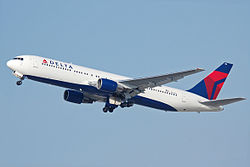}}} & \makecell{Q62161 \\ \imgborderred{\includegraphics[height=1.5cm]{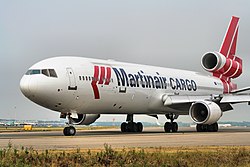}}} & \makecell{Q852 \\ \imgborderred{\includegraphics[height=1.5cm]{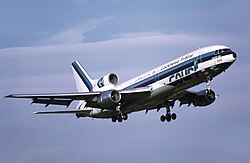}}} & \makecell{Q6423  \\ \includegraphics[height=1.5cm]{images/lead_images/Q6423.jpg}}\\
        
        \midrule
        
        \makecell{\includegraphics[height=1.5cm]{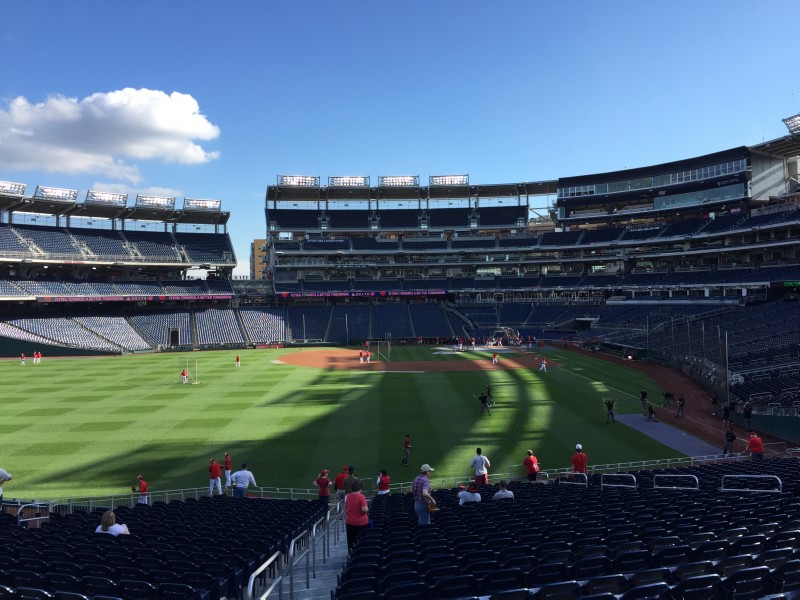}} & \makecell{what is the name of this park?}& \makecell{Q517545 \\ \imgbordergreen{\includegraphics[height=1.5cm]{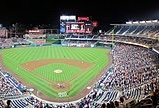}}} & \makecell{Q912770 \\ \imgborderred{\includegraphics[height=1.5cm]{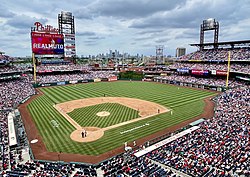}}} & \makecell{Q7381109 \\ \imgborderred{\includegraphics[height=1.5cm]{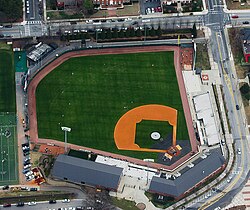}}} & \makecell{Q517545 \\ \includegraphics[height=1.5cm]{images/lead_images/Q517545.jpg}}\\
        
        \bottomrule
    \end{tabular}
    \caption{Example KnowCoL's predictions on five test images. For each query, the model’s top-3 (according to the cosine similarity) Wikidata items predictions are shown, with green borders indicating correct matches and red borders indicating incorrect ones; the ground-truth item is in the rightmost column.}
    \label{tab:Examples}
\end{table*}
To complement our quantitative evaluation, we present qualitative examples of KnowCoL’s predictions on the OVEN test set in Table~\ref{tab:Examples}. These cases illustrate the model’s ability to recognize fine-grained entities under zero-shot settings correctly.

Overall, KnowCoL can place the correct Wikidata Q-ID at rank 1 for most cases, even though those three are drawn from a pool of more than 30000 candidates.  Its main failure occurs when the top candidates are similar (e.g., different medieval ruins), causing the true entity to slip out of the top three entirely. We also observe that the top three candidates are all highly plausible alternatives to the ground truth.

\end{document}